\documentclass{article} 
\usepackage{iclr2027_conference,times}

\usepackage{amsmath,amsfonts,bm}

\def\eqref#1{equation~\ref{#1}}

\def\1{\bm{1}}

\DeclareMathAlphabet{\mathsfit}{\encodingdefault}{\sfdefault}{m}{sl}
\SetMathAlphabet{\mathsfit}{bold}{\encodingdefault}{\sfdefault}{bx}{n}

\usepackage{hyperref}
\usepackage{url}
\usepackage{amsmath}
\usepackage{graphicx}
\usepackage{subcaption}
\usepackage{booktabs}
\usepackage{algorithm}
\usepackage{algpseudocode}
\DeclareCaptionFont{slightly_smaller_subfig_font}{\fontsize{8.5pt}{9pt}\selectfont}
\usepackage{wrapfig}
\usepackage{amsthm}
\theoremstyle{plain}
\newtheorem{lemma}{Lemma}[section]
\usepackage[most]{tcolorbox}

\usepackage[most]{tcolorbox}

\definecolor{propaccent}{HTML}{315C7D}
\definecolor{propbackground}{HTML}{F4F7FA}
\definecolor{propborder}{HTML}{AABCCA}

\usepackage[most]{tcolorbox}

\definecolor{propblue}{HTML}{315D75}
\definecolor{propbackground}{HTML}{F4F7F9}

\newtcbtheorem[number within=section]
  {proposition}{Proposition}{
    enhanced,
    breakable,
    frame hidden,
    colback=black!3,
    coltitle=black,
    fonttitle=\bfseries,
    fontupper=\normalfont,
    sharp corners,
    borderline west={1.8pt}{0pt}{propaccent},
    attach title to upper,
    after title={\par\smallskip},
    left=8pt,
    right=8pt,
    top=6pt,
    bottom=6pt,
    before skip=8pt,
    after skip=8pt
  }{prop}

\newtcbtheorem[use counter from=proposition]
  {corollary}{Corollary}{
    enhanced,
    breakable,
    frame hidden,
    colback=black!3,
    coltitle=black,
    fonttitle=\bfseries,
    fontupper=\normalfont,
    sharp corners,
    borderline west={1.8pt}{0pt}{propaccent},
    attach title to upper,
    after title={\par\smallskip},
    left=8pt,
    right=8pt,
    top=6pt,
    bottom=6pt,
    before skip=8pt,
    after skip=8pt
  }{cor}

\title{Controlling Sharpening in LLM Post-Training}

\author{
Simon Matrenok\textsuperscript{1},
Nicolas Menet\textsuperscript{2},
Nicolas Flammarion\textsuperscript{1},
Andreas Krause\textsuperscript{2}
\\
\\
\textsuperscript{1}School of Computer and Communication Sciences (IC), EPFL\\
\textsuperscript{2}Department of Computer Science, ETH Zürich
}

\newcommand{\iid}{\mathrel{\overset{\mathrm{i.i.d.}}{\sim}}}

\iclrfinalcopy 
\begin{document}
\addtocontents{toc}{\protect\setcounter{tocdepth}{-1}}

\maketitle

\begin{abstract}
LLM post-training via reinforcement learning has shown consistent improvements in accuracy at the cost of reducing behavioral coverage---a phenomenon known as sharpening that can stall
progress on difficult tasks. Repeated-sampling objectives such as
$\mathrm{pass@k}$ and $\mathrm{max@k}$ are known to mitigate
sharpening, but the mechanism behind it remains poorly understood. Moreover, existing
approaches for $\mathrm{@k}$ optimization use dedicated algorithms, and no general
framework exists for adapting established
policy optimization methods from $\mathrm{@}1$ to $\mathrm{@k}$.
We derive the optimal policies for KL-regularized
$\mathrm{pass@k}$ and $\mathrm{max@k}$ objectives in closed form. Our explicit solutions reveal that $\mathrm{@k}$ optimization naturally redistributes learning pressure based on prompt difficulty, directing it toward harder problems. Crucially, we show that standard $\mathrm{@}1$ objectives achieve these exact same optimal policies through prompt-adaptive regularization or reward transformation. Ultimately, this yields an optimizer-independent explanation for why $\mathrm{@k}$ objectives reduce sharpening on easy tasks while prioritizing success on harder ones.
Building on these insights, we introduce {\em $\mathrm{@k}$ Adaptive Pressure for Policy Optimization (\textsc{KAPPO})}. Following a one-time reference calibration, \textsc{KAPPO} seamlessly adapts existing $\mathrm{@}1$ methods to $\mathrm{@k}$ via prompt-adaptive regularization or reward transformation, without otherwise altering their training procedures. Experiments with several RL methods on mathematical reasoning and code generation demonstrate that \textsc{KAPPO} provides direct control over sharpening. Crucially, we demonstrate that $\mathrm{@k}$ pressure can be fully realized with single-completion training, matching the performance of multi-completion $\mathrm{@k}$ optimizers at a fraction of the cost.\looseness-1
\end{abstract}

\section{Introduction}

Reinforcement learning (RL) has become a central component of LLM post-training, producing substantial performance improvements in mathematical reasoning, code generation, and other verifiable domains~\citep{shao2024deepseekmath,chen2025acereason,yu2025dapo}. Yet, an increase in average performance does not by itself reveal how the model has improved. Pretrained LLMs contain extensive latent capabilities, and RL can affect them through two distinct mechanisms.
It may uncover successful behavior that is rarely expressed by the base policy, or it may concentrate probability on solutions that are already readily available. The latter phenomenon, referred to as
\emph{sharpening}, improves single-trial success but often reduces the range of problems solvable by the model under repeated trials~\citep{yue2025reinforcement}. Further capability progress thus depends on effectively controlling sharpening.\looseness-1

Recent studies confirm this exact limitation: RL post-training dynamics are indeed dominated by sharpening~\citep{yue2025reinforcement,song2025outcome}. The impressive single-trial gains frequently diminish or reverse in multi-trial settings, as the generated reasoning paths are confined to a narrow subset of the base model's original distribution. Counteracting this over-concentration is notoriously difficult. For example, simply injecting randomness or diversity at the token level rarely translates into meaningful diversity in high-level reasoning strategies~\citep{walder2025pass, chen2025passktraining}.\looseness-1

Multi-trial metrics provide a practical lens for quantifying the extent of sharpening~\citep{yue2025reinforcement}. For binary rewards, $\mathrm{pass@k}$ measures the probability that at least one of $k$ completions succeeds; for continuous rewards, $\mathrm{max@k}$ measures the expected maximum reward among $k$ completions. Gains driven by sharpening concentrate at small $k$ and diminish as the sampling budget grows, whereas gains that persist at large $k$ indicate broader coverage of successful behavior. While these metrics do not prove that the model has discovered reasoning strategies outside its base distribution, they remain the clearest operational proxies for diverse reasoning. These observations motivate using $\mathrm{@k}$ metrics not just for evaluation, but also as training objectives to mitigate sharpening~\citep{walder2025pass}. However, the mechanism that makes $\mathrm{@k}$ objectives less prone to sharpening remains poorly understood. Moreover, existing approaches rely on dedicated algorithms for $\mathrm{@k}$ optimization, precluding a direct transfer of standard $\mathrm{@1}$ reinforcement learning advancements to the $\mathrm{@k}$ setting.\looseness-1

We derive the closed-form optimal policies for KL-regularized $\mathrm{pass@k}$ and $\mathrm{max@k}$. Our analysis reveals that $\mathrm{@k}$ optimization acts as prompt-adaptive pressure: it restricts optimization on tasks the model already solves reliably and redirects pressure toward
tasks on which successful behavior is rare. We show that this behavior is achievable with ordinary $\mathrm{@}1$ objectives via prompt-adaptive regularization or reward transformation.
Leveraging this connection, we introduce {\em $\mathrm{@k}$ Adaptive Pressure for Policy Optimization (\textsc{KAPPO})}, a framework for turning existing $\mathrm{@}1$ policy-optimization methods into
$\mathrm{@k}$ optimizers. After a one-time reference calibration, \textsc{KAPPO} introduces prompt-adaptive pressure while preserving all other characteristics of the base method. 
Our contributions are:
\begin{enumerate}

\item
We derive closed-form optimal policies for KL-regularized $\mathrm{pass@k}$ and $\mathrm{max@k}$ objectives.

\item
We prove that $\mathrm{@k}$ optimization redistributes optimization pressure according to task difficulty via prompt-adaptive regularization or reward transformation, prioritizing hard tasks.

\item
We introduce \textsc{KAPPO}, a framework that
turns existing $\mathrm{@}1$ methods into $\mathrm{@k}$ optimizers. In \textsc{KAPPO}, $k$ directly controls sharpening without requiring online groups of $k$ completions.

\item
We instantiate \textsc{KAPPO} across several policy optimization
methods, matching the performance of a direct multi-completion $\mathrm{@k}$-optimization baseline at a fraction of its rollouts.\looseness-1

\end{enumerate}

\section{Related Work}
\label{sec:related_work}

\paragraph{Exploration mechanisms in RLVR.}
Beyond the sharpening evidence discussed above, RL has been shown to consistently reduce generation diversity~\citep{kirk2024understanding},
including on problems the model has not yet
solved~\citep{song2025outcome}. Application of standard exploration strategies, such as entropy regularization~\citep{mnih2016asynchronous, haarnoja2018soft, ahmed2019understanding} and noisy reward perturbations~\citep{lu2017ensemble}, has shown limited or sensitive improvements relative to methods designed specifically for multi-attempt reasoning~\citep{walder2025pass,chen2025passktraining}. More structured approaches are guided by epistemic uncertainty via policy ensembles~\citep{menet2026poetsuncertaintyawarellmoptimization}, or introduce exploration bonuses based on final outcomes~\citep{song2025outcome}, hidden representations~\citep{tuyls2025representation, menet2026thompson}, or sequence-level diversity \citep{
hu2025diversity}. These methods rely on different proxies for novelty and have been evaluated in heterogeneous settings; no established exploration mechanism for LLM post-training has yet emerged.\looseness-1

\paragraph{\boldmath $\mathrm{Pass@k}$ and $\mathrm{max@k}$
optimization.}
Originally introduced as an evaluation metric for code
generation~\citep{chen2021evaluating}, $\mathrm{pass@k}$ has recently
been adopted as a training objective. \cite{walder2025pass} derive unbiased policy-gradient estimators, called PKPO, for binary $\mathrm{pass@k}$ and continuous
$\mathrm{max@k}$ rewards. They show that increasing $k$ during training preserves higher entropy, increases the number of problems solved at least once, and unblocks learning on difficult tasks where $\mathrm{pass@}1$ optimization stalls. $\mathrm{max@k}$ objectives are submodular~\citep{krause2014submodular}, and encourage ``hedging the bets'' given limited expressivity, a principle associated with exploration. Subsequent work studies analytical advantage design~\citep{chen2025passktraining}, token-level mechanisms for preventing probability over-concentration~\citep{peng2025simko}, improved policy-gradient estimators for order-statistic objectives~\citep{parmas2026ordergrad}, and a pass@k interpretation of maximum likelihood objectives~\citep{tajwar2026maximumlikelihoodreinforcementlearning}. These approaches primarily analyze training gradients or construct multi-completion learning signals. We instead derive the corresponding KL-regularized optimal policies and show that the effect of $\mathrm{@k}$ optimization can be reproduced through prompt-adaptive $\mathrm{@}1$ objectives. This provides a general route for turning existing $\mathrm{@}1$ methods into $\mathrm{@k}$ variants without constructing expensive $\mathrm{@k}$-dependent multi-completion learning signals.\looseness-1

\section{Preliminaries}

\paragraph{KL-regularized reward optimization.}
Let $x\sim\mathcal{D}$ denote a prompt, $y\sim\pi(\,\cdot\mid x)$ a
completion, and $R(y,x)$ the corresponding reward. Standard KL-regularized RL
optimizes the objective
\begin{equation}\label{eq:single_reward_objective}
\mathcal{J}(\pi)
:=
\mathbb{E}_{x\sim\mathcal{D}}
\left[
\mathbb{E}_{y\sim\pi(\cdot\mid x)}[R(y,x)]
-
\beta\,\mathrm{KL}
\left(
\pi(\, \cdot\mid x)\Vert\pi_{\mathrm{ref}}(\, \cdot\mid x)
\right)
\right],
\end{equation}
where $\pi_{\mathrm{ref}}$ is a fixed reference policy and $\beta>0$ controls the strength of regularization. While some recent RLVR methods omit the KL penalty to allow larger policy shifts~\citep{yu2025dapo}, other works suggest that carefully designed KL-regularized objectives enhance both reasoning performance and training stability~\citep{liu2025prorl}. Beyond potential empirical advantages, KL regularization is a prerequisite for non-degenerate optimal policies. Indeed, the optimal policy is given in closed form:\looseness-1
\begin{equation}\label{eq:optimal_policy}
\pi^*(y\mid x)
=
\tfrac{1}{Z(x)}
\pi_{\mathrm{ref}}(y\mid x)
\exp({R(y,x)}/{\beta}).
\end{equation}
Methods such as REINFORCE~\citep{williams1992simple}, RLOO~\citep{ahmadian2024back}, PPO~\citep{schulman2017ppo}, and GRPO~\citep{shao2024deepseekmath} approximate gradients of Eq.~\ref{eq:single_reward_objective}. In contrast, methods like DPO~\citep{rafailov2023direct}, $A^*$PO~\citep{brantley2025accelerating}, and QRPO~\citep{matrenok2026quantile} directly fit a parameterized policy $\pi_\theta$ to the optimal policy $\pi^*$ given in Eq.~\ref{eq:optimal_policy}. 
Our analysis and contributions are agnostic to the choice of optimizer, separating $\mathrm{@k}$ capabilities from other advancements.\looseness-1

\paragraph{\boldmath Pass@k and max@k.}
Let $y_1,\ldots,y_k \iid \pi(\,\cdot\mid x)$. We define $\mathrm{max@k}$ and $\mathrm{pass@k}$ as
\begin{equation*}
\mathrm{max@k}(\pi;x) = \mathbb{E}
\big[
\max_{i=1,\ldots,k}R(y_i,x)
\big], \qquad \mathrm{pass@k}(\pi;x)
=
\mathbb{P}\left[\exists\, i \in \{1,\ldots, k\} : R(y_i,x)=1\right].
\end{equation*}
Note that for binary rewards $R(y,x)\in\{0,1\}$, $\mathrm{max@k}$ coincides with
$\mathrm{pass@k}$. Moreover, for $k=1$, both expressions recover the ordinary $\mathrm{@1}$ expected reward $\mathbb{E}_{y\sim\pi(\cdot\mid x)}[R(y,x)]$ used in the standard RL objective $\mathcal J$ in Eq.~\ref{eq:single_reward_objective}. Let $V_k(\pi;x)$ denote either of these $\mathrm{@k}$ objectives. Substituting $V_k(\pi;x)$ into $\mathcal J$ in place of the $\mathrm{@1}$ reward yields the KL-regularized objective studied in this work:\looseness-1
\begin{equation}\label{eq:k_sample_objective}
\mathcal{J}_k(\pi)
=
\mathbb{E}_{x\sim\mathcal{D}}
\left[
V_k(\pi;x)
-
\beta\,\mathrm{KL}
\left(
\pi(\, \cdot\mid x)\Vert\pi_{\mathrm{ref}}(\, \cdot\mid x)
\right)
\right].
\end{equation}

\section{$\mathrm{@k}$ Optimal Policies and Adaptive Pressure}\label{sec:optimal_policies}
\subsection{Binary Rewards:
\texorpdfstring{$\mathrm{pass@k}$}{pass@k}}

\begin{figure}[t]
    \centering
    \begin{subfigure}[t]{0.31\linewidth}
        \centering
        \includegraphics[width=\linewidth]
        {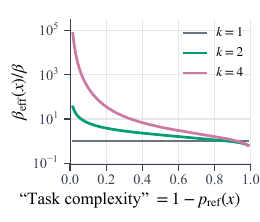}
        \caption{$\mathrm{pass@k}$: effective regularization}
        \label{fig:passk_effective_beta}
    \end{subfigure}\hfill
    \begin{subfigure}[t]{0.31\linewidth}
        \centering
        \includegraphics[width=\linewidth]
        {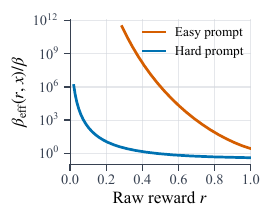}
        \caption{$\mathrm{max@k}$: effective regularization}
        \label{fig:maxk_effective_beta}
    \end{subfigure}\hfill
    \begin{subfigure}[t]{0.31\linewidth}
        \centering
        \includegraphics[width=\linewidth]
        {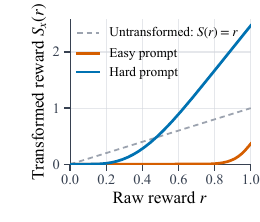}
        \caption{$\mathrm{max@k}$: reward transformation}
        \label{fig:maxk_reward_transform}
    \end{subfigure}

    \caption{
    Adaptive pressure induced by $\mathrm{@k}$ objectives.
    \textbf{(a)} For binary rewards, increasing $k$ strengthens
    regularization on easy tasks while retaining greater optimization
    pressure on difficult tasks.
    \textbf{(b--c)} For continuous rewards, we illustrate the same
    effect at $k=4$ using synthetic ``easy'' $\mathrm{Beta}(8,2)$ and
    ``difficult'' $\mathrm{Beta}(2,8)$ reference reward distributions on
    $[0,1]$. The easy task receives stronger effective regularization
    and smaller transformed rewards than the difficult task. All panels
    use $\beta=0.5$.
    }
    \label{fig:adaptive_pressure}
\end{figure}

Proofs of all results in this subsection, including boundary cases,
are provided in Appendix~\ref{app:passk_derivation}.
We first consider binary rewards, i.e., we set $V_k(\pi;x) = \mathrm{pass@k}(\pi;x)$ in
Eq.~\ref{eq:k_sample_objective} for $\mathcal J_k$, and derive the closed-form optimal policy which maximizes $\mathcal J_k$. It takes the familiar exponential-tilt form.
\begin{proposition}{Closed-form optimal policy for the \boldmath $\mathrm{pass@k}$ objective}{passk_optimal_policy}
    Let $R(y,x)\in\{0,1\}$. Then the optimal policy for $\mathcal J_k$ is
    \begin{equation}\label{eq:passk_optimal_policy}
    \pi^*(y\mid x)
    =
    \frac{1}{Z(x)}
    \pi_{\mathrm{ref}}(y\mid x)
    \exp\left(
    \frac{R(y,x)}{\beta_{\mathrm{eff}}(x)}
    \right),
    \end{equation}
    \begin{equation}\label{eq:passk_effective_beta}
    \text{where } \beta_{\mathrm{eff}}(x)
    =
    \frac{\beta}{k(1-p^*(x))^{k-1}}
    \text{ and }
    Z(x)
    =
    1-p_{\mathrm{ref}}(x)
    +
    p_{\mathrm{ref}}(x)
    \exp\left(\frac{1}{\beta_{\mathrm{eff}}(x)}\right).
    \end{equation}
    Here,
    $p_{\mathrm{ref}}(x)
    =\mathbb P_{y\sim\pi_{\mathrm{ref}}(\cdot\mid x)}[R(y,x)=1]$
    and
    $p^*(x)
    =\mathbb P_{y\sim\pi^*(\cdot\mid x)}[R(y,x)=1]$
    are, respectively, the success probabilities of the reference model and optimal model.
\end{proposition}
Compared with the optimal policy (Eq.~\ref{eq:optimal_policy}) for the single-reward objective (Eq.~\ref{eq:single_reward_objective}), the constant regularization strength $\beta$ is replaced by the prompt-dependent effective regularization strength $\beta_{\mathrm{eff}}(x)$.
The obtained expression for the optimal policy, however, remains implicit since $p^*(x)$ depends on the unknown optimal policy $\pi^*$. Nevertheless, it can be made fully explicit by summing Eq.~\ref{eq:passk_optimal_policy} over successful completions, which yields a scalar equation relating $p^*(x)$ directly to $p_{\mathrm{ref}}(x)$.
\begin{proposition}
{Scalar fixed-point characterization}
{passk_fixed_point}
For $0<p_{\mathrm{ref}}(x)<1$, the optimal success probability
$p^*(x)$ is the unique solution in $(0,1)$ of
\begin{equation}\label{eq:passk_fixed_point}
k(1-p^*(x))^{k-1}
=
\beta\log
\frac{p^*(x)(1-p_{\mathrm{ref}}(x))}
{p_{\mathrm{ref}}(x)(1-p^*(x))}.
\end{equation}
\end{proposition}
The reference success probability $p_{\mathrm{ref}}(x)$ can be estimated directly from reference-model samples,\footnote{The same reference samples can be reused across training epochs and hyperparameter sweeps.}, while $p^*(x)$ is computed by solving Eq.~\ref{eq:passk_fixed_point}. For an error analysis, see Appendix~\ref{app:calibration_sensitivity}. Proposition~\ref{prop:passk_optimal_policy} suggests two equivalent ways to recover the $\mathrm{pass@k}$ optimal policy with single-reward optimization.\looseness-1
\begin{corollary}
  {\boldmath $\mathrm{pass@k}$ as prompt-adaptive regularization}  {passk_adaptive_regularization}
Optimizing the ordinary single-reward objective in
Eq.~\ref{eq:single_reward_objective} using the prompt-adaptive
regularization strength $\beta_{\mathrm{eff}}(x)$ from
Eq.~\ref{eq:passk_effective_beta} results exactly in the
$\mathrm{pass@k}$ optimal policy provided in
Eq.~\ref{eq:passk_optimal_policy}.\looseness-1
\end{corollary}
This equivalence holds because both objectives decompose over prompts
and $\beta_{\mathrm{eff}}(x)$ is constant w.r.t. the
completion $y$ for each fixed $x$. As the optimal policy given in Eq.~\ref{eq:passk_optimal_policy} depends on the reward and
KL coefficient only through their ratio, we obtain a
complementary reward-side interpretation.\looseness-1
\begin{corollary}
  {\boldmath $\mathrm{pass@k}$ as prompt-adaptive reward scaling}
  {passk_adaptive_reward}
Optimizing the ordinary single-reward objective $\mathcal J$ given in
Eq.~\ref{eq:single_reward_objective} with constant regularization
strength $\beta$ but applying a prompt-adaptive reward transformation
\begin{equation}\label{eq:passk_adaptive_reward}
\widetilde R_k(y,x)
=
\tfrac{\beta}{\beta_{\mathrm{eff}}(x)}R(y,x)
=
k(1-p^*(x))^{k-1}R(y,x)
\end{equation}
results exactly in the $\mathrm{pass@k}$ optimal policy provided in
Eq.~\ref{eq:passk_optimal_policy}.
\end{corollary}
\paragraph{Two views of adaptive pressure.}
Corollary~\ref{cor:passk_adaptive_regularization} and Corollary~\ref{cor:passk_adaptive_reward} show that $\mathrm{pass@k}$ optimization can be expressed as ordinary single-reward optimization with either prompt-adaptive regularization or, equivalently, prompt-adaptive reward scaling. Taking $1-p_{\mathrm{ref}}(x)$ as a simple measure of prompt complexity, easy tasks $x$ have high $p_{\mathrm{ref}}(x)$ and consequently high $p^*(x)$, resulting in stronger effective regularization, while hard tasks receive weaker effective regularization and therefore stronger optimization pressure (see Figure~\ref{fig:passk_effective_beta}). Equivalently, under the reward-side view, a successful completion receives a smaller reward multiplier on easy tasks and a larger multiplier on difficult ones.

\paragraph{Intuition: reducing sharpening.}
Consider optimizing the usual single-reward objective (Eq.~\ref{eq:single_reward_objective}) on a binary-reward task. The reward gain depends only on the increase in success probability, regardless of initial performance---for example, moving from near zero to $0.1$ on a hard task is worth as much as moving from $0.9$ to $1$ on an already-easy task. For an LLM, however, the latter is often much easier: the model already knows how to solve the task and only needs to concentrate probability on known successful behavior. 
In contrast, increasing a near-zero success probability requires acquiring entirely novel reasoning strategies. The adaptive pressure induced by $\mathrm{pass@k}$ objectives counteracts this tendency towards sharpening by imposing stronger effective regularization on already-easy tasks or, equivalently, through the prompt-adaptive reward scaling described above.\looseness-1

\newpage

\subsection{Continuous Rewards:
\texorpdfstring{$\mathrm{max@k}$}{max@k}}
Next, we consider real-valued rewards, i.e., we set $V_k(\pi;x) = \mathrm{max@k}(\pi;x)$ in
Eq.~\ref{eq:k_sample_objective} for $\mathcal J_k$, and derive the associated closed-form optimal policy which maximizes $\mathcal J_k$. For $k > 1$, we obtain:
\begin{proposition}{\boldmath Closed-form optimal policy for the $\mathrm{max@k}$ objective}{maxk_adaptive_reward}
The optimal policy for the $\mathrm{max@k}$ objective $\mathcal J_k$ under real-valued rewards evaluates to
\begin{equation}\label{eq:maxk_optimal_policy_implicit}
\pi^*(y\mid x)
=
\frac{1}{Z(x)}
\pi_{\mathrm{ref}}(y\mid x)
\exp\left(
\frac{k}{\beta}
\int_{-\infty}^{R(y,x)}
F_{\pi^*}(r\mid x)^{k-1}\,dr
\right),
\end{equation}
where
$F_{\pi^*}(r\mid x)
:=
\mathbb{P}_{y'\sim\pi^*(\cdot\mid x)}
[R(y',x)\leq r]$
is the reward CDF under the optimal policy.
\end{proposition}
The optimal policy identified above remains implicit because the CDF $F_{\pi^*}$ on its right-hand side is itself induced by the unknown optimal policy $\pi^*$. In contrast to the binary-reward case, computing $F_{\pi^*}$ is more difficult and requires a sufficiently accurate estimate of the entire reference reward distribution $F_{\pi_{ref}}$. It can, nevertheless, be computed numerically before training; we provide the procedure in Appendix~\ref{app:maxk_reconstruction}. All derivations and regularity conditions are explained in Appendix~\ref{app:maxk_derivation}.

Following the derivation for $\mathrm{pass@k}$, a natural question is whether we can also identify a task-dependent $\beta_{\mathrm{eff}}$ here that reduces $\mathrm{@k}$ optimization to $\mathrm{@1}$. A simple substitution shows that:\looseness-1
\begin{equation}\label{eq:maxk_effective_beta}
\pi^*(y\mid x)
=
\tfrac{1}{Z(x)}
\pi_{\mathrm{ref}}(y\mid x)
\exp\left(
\tfrac{R(y,x)}
{\beta_{\mathrm{eff}}(R(y,x),x)}
\right) \text{ for }\,
\beta_{\mathrm{eff}}(r,x)
= \frac{\beta}{k}  r\Big/\!\!
{\displaystyle\int_{-\infty}^r\!\!\!\!\!F_{\pi^*}(u\mid x)^{k-1}\,du}.
\end{equation}
Note that, to ensure well-definedness, one can assume $R(y,x) > 0$ without loss of generality given (lower) bounded rewards, since the objectives in Eqs.~\ref{eq:single_reward_objective} and~\ref{eq:k_sample_objective} are invariant to constant reward shifts.
%

Eq.~\ref{eq:maxk_effective_beta} provides an interpretation analogous to the binary-reward case:
$\mathrm{max@k}$ replaces constant regularization with a strength that adapts to task complexity. In the $\mathrm{max@k}$ case, however, the effective strength also varies with the reward of the individual completion. As an illustrative example, Figure~\ref{fig:maxk_effective_beta} uses two synthetic reference reward distributions on the $[0,1]$ interval representing an ``easy'' and a ``difficult'' task. This completion dependence means that substituting $\beta_{\mathrm{eff}}(R(y,x),x)$ into the ordinary single-reward objective would produce a different
optimal policy; hence, the exact adaptive-regularization equivalence established in Corollary~\ref{cor:passk_adaptive_regularization} for binary rewards does not extend to continuous rewards. Still, $\beta_{\mathrm{eff}}$ provides a useful interpretation of how $\mathrm{max@k}$ redistributes optimization pressure and could be adapted in practice to design completion-adaptive regularization.
%
Alternatively, we may follow our derivations for $\mathrm{pass@k}$ and instead consider adaptive reward transformations to redistribute optimization pressure. We obtain the following objective-level equivalence:\looseness-1

\begin{corollary}
  {\boldmath $\mathrm{max@k}$ as prompt-adaptive reward transformation}
  {maxk_adaptive_reward}
Optimizing the ordinary single-reward objective $\mathcal J$ given in Eq.~\ref{eq:single_reward_objective} with the transformed reward
\begin{equation}\label{eq:maxk_reward_transform}
\widetilde R(y,x)=S_x(R(y,x)) \,\text{ for }\, S_x(r)
=
k\int_{-\infty}^r
F_{\pi^*}(u\mid x)^{k-1}\,du
\end{equation}
results in the $\mathrm{max@k}$ optimal policy derived in Proposition~\ref{prop:maxk_adaptive_reward}.
\end{corollary}
The transformed-reward view shows the same prompt-adaptive effect. In the illustration in Figure~\ref{fig:maxk_reward_transform}, the same raw reward is assigned a larger transformed value on the difficult task than on the easy task, providing a complementary reward-space view of how $\mathrm{max@k}$ can be used to control sharpening.

\paragraph{Explicit closed-form optimal policy for win-rate reward.} 
As discussed, the optimal policy in Eq.~\ref{eq:maxk_optimal_policy_implicit} remains implicit because it depends on the unknown CDF $F_{\pi^*}$. A workaround is to consider a different but closely related reward: the reference quantile $F_{\mathrm{ref}}(r\mid x) =\mathbb{P}_{y'\sim\pi_{\mathrm{ref}}(\cdot\mid x)}[R(y',x)\leq r]$, which preserves reward ordering and represents the win rate against reference-policy completions. Indeed, the $\mathrm{max@}k$ optimal policy for these win-rate rewards admits a fully explicit expression.
\begin{proposition}
{\boldmath Explicit optimal policy for win-rate $\mathrm{max@}k$}
{winrate_maxk_optimal_policy}
For a continuous reference reward distribution with CDF $F_{\mathrm{ref}}(r \mid x)$, the objective $\mathcal J_k$ with win-rate rewards $F_{\mathrm{ref}}(R(y,x) \mid x)$ admits the following optimal policy:
\begin{equation}\label{eq:quantile_maxk_optimal_policy}
\pi_q^*(y\mid x)
=
\pi_{\mathrm{ref}}(y\mid x)
\exp\left(\frac{S_x^{\mathrm q}(R(y,x))}{\beta}\right),
\end{equation}
where
\begin{equation}\label{eq:quantile_maxk_transform}
S_x^{\mathrm{q}}(r)
=
\beta\log\big(
C+\frac{1}{\beta}
\Psi^{-1}\!\left(F_{\mathrm{ref}}(r\mid x)\right)^k
\!\big),\, \Psi(z)
\!=
\!\!\int_0^z\!\!\!
\frac{du}{C+\frac{1}{\beta}u^k},\text{ and } C \text{ s.t. } \Psi(1) = 1.
\end{equation}
\end{proposition}
For a proof, see Appendix~\ref{app:winrate_maxk_derivation}.
The scalar $C$ and mapping $\Psi$ depend only on
$k$ and $\beta$, and can be numerically pre-computed prior to training. The only prompt-dependent quantity in $S_x^{\mathrm{q}}$ is $F_{\mathrm{ref}}(r\mid x)$, which,
similarly to $p_{\mathrm{ref}}(x)$ for binary-rewards, can be
estimated once from a fixed bank of reference-model completions (see Appendix~\ref{app:calibration_sensitivity} for an error analysis).
Thus, the adaptive-reward principle of Corollary~\ref{cor:maxk_adaptive_reward} admits a directly computable realization for win-rate-based optimal policies.\looseness-1

\section{\textsc{KAPPO}:
\texorpdfstring{$\mathrm{@k}$}{@k} Adaptive Pressure for Policy
Optimization}
\label{sec:kappo}

We introduce \emph{$\mathrm{@k}$ Adaptive Pressure for Policy Optimization} (\textsc{KAPPO}), a framework for adapting existing $\mathrm{@}1$ policy-optimization methods to $\mathrm{@k}$ settings. The parameter $k$ can be tuned directly to control sharpening without otherwise changing the optimization algorithm. Given a base optimizer, \textsc{KAPPO} redistributes optimization pressure either through adaptive reward transformation or adaptive regularization, following our derivations in Section~\ref{sec:optimal_policies}. The required prompt-dependent transformations and coefficients are estimated once before training at negligible cost and
reused across runs; we call this stage \emph{reference calibration} (see Appendix~\ref{app:reference_calibration}). 
We instantiate both interfaces below to demonstrate  \textsc{KAPPO}'s generality and effectiveness across several popular RLVR optimizers.\looseness-1

\subsection{Regularization-Side Adaptation:
\texorpdfstring{$\mathrm{@k}$-DPO}{@k-DPO}}

The regularization-side interface is particularly useful for methods that do not directly optimize scalar rewards, most prominently DPO~\citep{rafailov2023direct}. We apply it in the binary-reward setting,\footnote{We note that prompt-adaptive regularization is also possible for continuous rewards, but without the same objective-level equivalence, e.g., by directly adjusting the regularization coefficient of the gradient expression.} using the connection established in
Corollary~\ref{cor:passk_adaptive_regularization} to replace DPO's global KL coefficient with the prompt-dependent $\beta_{\mathrm{eff}}(x)$ from Eq.~\ref{eq:passk_effective_beta}. We call the resulting method $\mathrm{@k}$-DPO:\looseness-1 
\begin{equation}\label{eq:passk_dpo_loss}
\mathcal{L}_{\mathrm{@k}\text{-}\mathrm{DPO}}(\theta)
=
-\mathbb{E}_{(x,y_w,y_l)\sim\mathcal{D}_{\mathrm{pref}}}
\log\sigma\left(
\beta_{\mathrm{eff}}(x)
\left[
\log\tfrac{\pi_\theta(y_w\mid x)}
         {\pi_{\mathrm{ref}}(y_w\mid x)}
-
\log\tfrac{\pi_\theta(y_l\mid x)}
         {\pi_{\mathrm{ref}}(y_l\mid x)}
\right]
\right).
\end{equation}
Before training, we perform reference calibration by generating a fixed bank of $M$ reference-model completions for each prompt $x$ and use the associated binary rewards to estimate $p_{\mathrm{ref}}(x)$. We then solve Eq.~\ref{eq:passk_fixed_point} for $p^*(x)$ and use it to compute $\beta_{\mathrm{eff}}(x)$. Each online update subsequently requires only a pair of completions per prompt. Except for $\beta_{\mathrm{eff}}$, the DPO training procedure remains unchanged.

\subsection{Reward-Side Adaptation:
\texorpdfstring{$\mathrm{@k}$-$A^{\star}$PO}{@k-A*PO} and \texorpdfstring{$\mathrm{@k}$-GRPO}{@k-GRPO}}
To demonstrate the reward-side interface, we consider two complementary instantiations. First, we choose $A^{\star}$PO~\citep{brantley2025accelerating}, a method that uses only a single online completion per prompt. This choice isolates a central implication of our analysis: online tuples of $k$ completions are not necessary to induce $\mathrm{@k}$ optimization pressure. Second, we also apply the interface to GRPO~\citep{shao2024deepseekmath}. Following common practice in RLVR, we use GRPO without a KL regularization term~\citep{yu2025dapo}; this additionally tests whether \textsc{KAPPO} remains effective beyond the KL-regularized setting assumed by our theory.

\textbf{\boldmath $\mathrm{@k}$-$A^{\star}$PO.}
$A^{\star}$PO optimizes the squared-regression loss
\begin{equation}\label{eq:astarpo_loss}
\mathcal{L}_{A^{\star}\mathrm{PO}}(\theta)
=
\mathbb{E}_{
x\sim\mathcal{D},\,
y\sim\pi_{\mathrm{old}}(\cdot\mid x)
}
\left[\big(
R(y,x)
-
\beta\log Z(x)
-
\beta\log
\tfrac{\pi_\theta(y\mid x)}
     {\pi_{\mathrm{ref}}(y\mid x)}
\big)^2\right],
\end{equation}
with the partition function $Z(x)$ estimated from reference-model completions. Replacing $R(y,x)$ with the prompt-adaptive reward yields $\mathrm{@k}$-$A^{\star}$PO. We treat both binary and continuous rewards:

\textit{For binary rewards}, we substitute the adaptive reward $\tfrac{\beta R(y,x)}{\beta_{\mathrm{eff}}(x)}$ into Eq.~\ref{eq:astarpo_loss}. We obtain $\beta_{\mathrm{eff}}$ and $Z(x)$ through the same one-time reference calibration procedure as for the binary $\mathrm{@k}$-DPO instantiation.
    
\textit{For continuous rewards}, we apply the win-rate-based adaptive reward transformation $S_x^{\mathrm{q}}$ from Eq.~\ref{eq:quantile_maxk_transform} and set $Z(x) = 1$ since normalization follows automatically (see Proposition~\ref{prop:winrate_maxk_optimal_policy}). We conduct a one-time reference calibration to approximate $F_{\mathrm{ref}}(r \mid x)$. As in the binary case, we sample a fixed bank $\{y_j^{\mathrm{ref}}\}_{j=1}^{M}$ of reference-model completions for each prompt. We then use this bank to construct the approximation $\widehat F_{\mathrm{ref}}(r \mid x) = \frac{1}{M}\sum_{j=1}^{M} \mathbb{I}[R(y_j^{\mathrm{ref}},x)\leq r]$ for each prompt $x$.\looseness-1

\textbf{\boldmath $\mathrm{@k}$-GRPO.} 
Here, we construct $\mathrm{@k}$-GRPO for the continuous-reward setting. For the corresponding binary-reward adaptation as well as additional properties of GRPO instantiations, see Appendix~\ref{app:grpo_adaptation}.  Given a group of $n$ completions with rewards $R_i=R(y_i,x)$, we replace the raw rewards by the win-rate-based transformed rewards $S_x^{\mathrm q}(R_i)$ in the GRPO loss:
\begin{gather}\label{eq:kappo_grpo_loss}
\textstyle \mathcal{L}_{\mathrm{@k}\text{-}\mathrm{GRPO}}(\theta) = 
-\mathbb{E}_{
x\sim\mathcal{D},\,
y_1, \ldots, y_n \sim \pi_{\mathrm{old}}(\cdot\mid x)
}
\big[
\frac{1}{n}
\sum_{i=1}^{n}
\frac{1}{|y_i|}
\sum_t
u^{\mathrm{clip}}_{i,t}(\theta)
\big],
\\\nonumber
u^{\mathrm{clip}}_{i,t}(\theta) \!=\! \min\!\big(
\tfrac{\pi_\theta(y_i^t|y_i^{<t}\!\!,x)}{\pi_{\mathrm{old}}(y_i^t|y_i^{<t}\!\!,x)}\widehat A_i^{k},
\operatorname{clip}\big(
\tfrac{\pi_\theta(y_i^t|y_i^{<t}\!\!,x)}{\pi_{\mathrm{old}}(y_i^t|y_i^{<t}\!\!,x)},1{-}\epsilon,1{+}\epsilon
\big)\widehat A_i^{k}
\big), \ \widehat A_i^{k}
\!=\!
\tfrac{
S_x^{\mathrm q}(R_i)
-
\frac{1}{n}\!\sum_{j=1}^{n}\! S_x^{\mathrm q}(R_j)
}{
\operatorname{std}_{j=1,\ldots,n}
\left[S_x^{\mathrm q}(R_j)\right]
}.
\end{gather}
$\mathrm{@k}$-GRPO uses the same one-time reference calibration as the continuous $\mathrm{@k}$-$A^{\star}$PO instantiation. Apart from the reward transformation $S_x^\mathrm{q}$, the standard GRPO procedure remains unchanged.

\section{Experiments}
\label{sec:experiments}

\subsection{Setup}

\paragraph{Tasks and scope.}
We evaluate \textsc{KAPPO} on mathematical reasoning with binary rewards (correctness) and code generation with continuous rewards (test case pass rate). We investigate all instantiations of \textsc{KAPPO} presented in Section~\ref{sec:kappo}, but note that \textsc{KAPPO} could be applied to other algorithms.\looseness-1

\paragraph{Model selection.}
Two principles guide our choice of initial checkpoints. First, we avoid models that have undergone reasoning-focused preference optimization or RL. This reflects the practical setting in which RL for reasoning is applied directly after supervised fine-tuning (SFT), and separates the effect of our training from reasoning behavior acquired via an undocumented prior curriculum. Second, we aim to minimize possible contamination of the training and evaluation datasets. These criteria lead us to \texttt{OLMo-3-7B-Instruct-SFT} and \texttt{Qwen3-4B-Base}. The former is an SFT-only, pre-RL checkpoint with a fully open training pipeline including explicit data decontamination procedures~\citep{olmo2025olmo3}. For a second model family, we use the Qwen pre-trained rather than a post-trained checkpoint to limit exposure to undocumented post-training data, although its pretraining corpus cannot be fully audited~\citep{yang2025qwen3}. Hyperparameters and checkpoints are selected on a validation split; complete experimental details are provided in Appendix~\ref{app:experimental_details}.

\paragraph{Baselines.}
As elaborated in Section~\ref{sec:kappo}, \textsc{KAPPO} can be applied to a wide range of RL optimizers by adjusting the effective regularization or transforming the reward. Since applying \textsc{KAPPO} with $k=1$ leaves the underlying  $\mathrm{@1}$ optimizer unchanged, we use $k=1$ as the primary baseline to isolate the effect of adaptive $\mathrm{@k}$ pressure. 
We also compare a \textsc{KAPPO}-adjusted $A^*$PO algorithm with PKPO~\citep{walder2025pass}, a dedicated $\mathrm{pass@k}$ optimization algorithm that requires at least $n\geq k$ new on-policy completions per prompt.\looseness-1

\subsection{Binary Rewards: Mathematical Reasoning}
\label{sec:math_experiments}

\begin{figure}[b]
    \centering
    \begin{subfigure}[t]{0.492\linewidth}
        \centering
        \includegraphics[width=\linewidth]
        {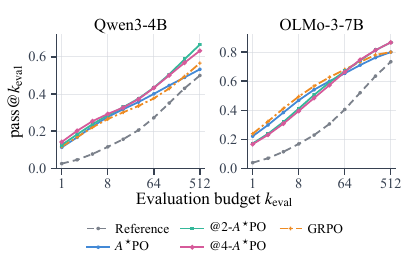}
        \caption{$A^{\star}$PO and $\mathrm{@k}$-$A^{\star}$PO.}
        \label{fig:aime26_kappo_grpo}
    \end{subfigure}\hfill
    \begin{subfigure}[t]{0.492\linewidth}
        \centering
        \includegraphics[width=\linewidth]
        {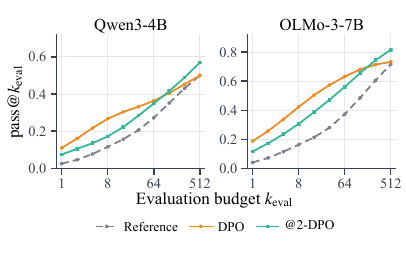}
        \caption{DPO and $\mathrm{@k}$-DPO.}
        \label{fig:aime26_dpo}
    \end{subfigure}
    \caption{
    $\mathrm{pass@k}_{\mathrm{eval}}$ on AIME 2026 as the
    sampling budget at test time ($\mathrm{k}_\mathrm{eval}$) increases. Setting $k>1$ during training consistently reduces sharpening. In Figure
    \textbf{(a)}, we plot standard GRPO and $\mathrm{@k}$-$A^{\star}$PO trained with
    $k\in\{1,2,4\}$. In Figure 
    \textbf{(b)}, we plot $\mathrm{@k}$-DPO trained with
    $k\in\{1,2\}$.\looseness-1
    }
    \label{fig:math_aime_passk}
\end{figure}

\begin{figure}[t]
    \centering
    \begin{subfigure}[t]{0.492\linewidth}
        \centering
        \includegraphics[width=\linewidth]
        {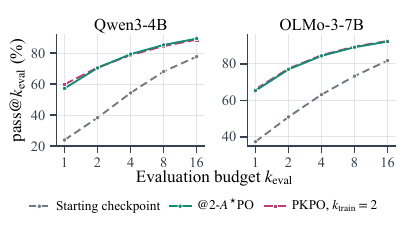}
        \label{fig:math_pkpo_k2}
    \end{subfigure}\hfill
    \begin{subfigure}[t]{0.492\linewidth}
        \centering
        \includegraphics[width=\linewidth]
        {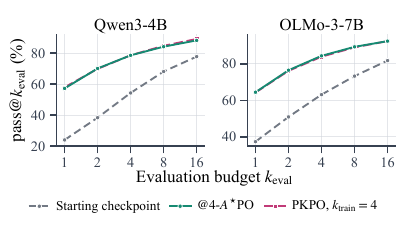}
        \label{fig:math_pkpo_k4}
    \end{subfigure}
    \vspace{-2.6\baselineskip}
    \caption{
    MATH Level~5 validation performance of the starting checkpoints, $\mathrm{@}k$-$A^{\star}$PO checkpoints, and PKPO checkpoints. $\mathrm{@}k$-$A^{\star}$PO matches the performance of PKPO at a fraction of the on-policy training rollouts, requiring only a single-time reference calibration at $\leq5\%$ training time overhead.
    }
    \label{fig:math_pkpo_comparison}
\end{figure}

\paragraph{Data and evaluation.}
We design the train--validation--test split to mirror a common model development setting: training data and validation benchmarks are closely coupled during development, while the final test consists of newer and often more difficult problems, appearing after model release. To closely match this setting, we perform RL on the training split of MATH and use its test split for in-distribution validation~\citep{hendrycks2021measuring}. Final evaluation is performed on AIME 2026~\citep{maa2026aime}, a markedly harder dataset whose problems predominantly exceed MATH's highest difficulty tier. AIME 2026 also postdates both model initialization checkpoints, ruling out direct contamination of our test benchmark. A completion receives reward one if its extracted final answer is correct and zero otherwise, i.e., we consider binary rewards.\looseness-1

\paragraph{Controlling sharpening by adjusting $k$.}
Figure~\ref{fig:math_aime_passk} plots $\mathrm{pass@k}_{\mathrm{eval}}$ against the number of completions $k_{\mathrm{eval}}$. We examine the improvement over the initial checkpoint as a function of the sampling budget. Gains \textit{exclusively} at small $k_{\mathrm{eval}}$ indicate that training has further concentrated probability on already-accessible behavior, whereas gains at large $k_{\mathrm{eval}}$ point toward acquisition of novel behavior. Figure~\ref{fig:aime26_kappo_grpo} considers reward-side adaptation. The $A^{\star}$PO baseline ($\mathrm{pass@1}$ training) closely tracks GRPO, showing that single-completion optimization performs competitive to multi-completion optimization for $\mathrm{pass@1}$ training. However, the gains achieved for both models diminish as $k_{\mathrm{eval}}$ increases. Training with $k>1$ typically (but not always) retains less of the $\mathrm{pass@}1$ gain achieved by $k=1$ training, but consistently overtakes the single-reward baselines at larger evaluation budgets. Figure~\ref{fig:aime26_dpo} exhibits the same effect for the regularization-side interface. Standard DPO produces gains concentrated at small evaluation budgets, whereas $\mathrm{@k}$-DPO trained with $k=2$ overtakes it as $k_{\mathrm{eval}}$ increases. This pattern holds for both OLMo and Qwen. 
Together, these results corroborate the effect predicted by our analysis across two base optimizers and both \textsc{KAPPO} interfaces: $\mathrm{@k}$ adaptation counteracts excessive sharpening and shifts optimization pressure from already-easy tasks toward difficult tasks where successful behavior remains rare.

\paragraph{Agreement with dedicated multi-completion $\mathrm{pass@k}$ optimization.}
We next verify that a \textsc{KAPPO} adaptation of a single-completion optimizer can match the performance of a state-of-the-art dedicated multi-completion $\mathrm{pass@k}$ algorithm. To that end, we compare $\mathrm{@k}$-$A^{\star}$PO with PKPO for $k\in\{2,4\}$ using a matched number of prompt batches and the minimum feasible PKPO rollout multiplicity ($n=k$). Across both models, $\mathrm{@k}$-$A^{\star}$PO achieves validation performance comparable to PKPO (see Figure~\ref{fig:math_pkpo_comparison}) while requiring only a fraction of the on-policy rollouts, reducing training cost. This supports our central prediction: $\mathrm{@k}$ optimization can be achieved without relying on online $k$-completion learning signals, but by injecting adaptive pressure into single-completion optimizers.\looseness-1


\subsection{Continuous Rewards: Code Generation}
\label{sec:code_experiments}

\paragraph{Data and evaluation.}
We follow the same data selection strategy as in the binary-reward
experiments: training and validation use closely related data, while
final evaluation uses a newer and substantially harder problem source
to approximate the distribution shift encountered after model
deployment. We conduct our $\mathrm{max@k}$ experiments using
\texttt{Qwen3-4B-Base}. Specifically, we train on the training split of
LeetCodeDataset~\citep{xia2025leetcode} and use its test split for validation. The two splits
share the same source and task format, with a temporal cutoff on July
2024~\citep{xia2025leetcode}. Each completion receives a continuous
reward in $[0,1]$ equal to the fraction of test cases it passes.
For final evaluation, we use the hard subset of LiveCodeBench v6, a substantially harder collection of
competitive-programming problems~\citep{jain2024livecodebench} released between January and April 2025, limiting the opportunity for their inclusion in
Qwen3 pretraining (released April 28th). Nevertheless, since Qwen's pretraining cutoff is not
documented, contamination cannot be ruled out completely.\looseness-1

\begin{wrapfigure}[18]{r}{0.50\linewidth}
    \vspace{-1.66\baselineskip}
    \centering
    \includegraphics[width=\linewidth]
    {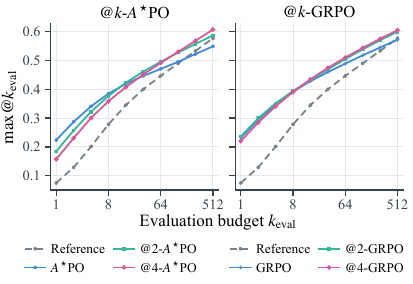}
    \vspace{-0.85\baselineskip}
    \caption{$\mathrm{max@}k_{\mathrm{eval}}$ on the hard subset of LiveCodeBench v6 for Qwen3-4B.
    Win-rate-based $\mathrm{@}k$-$A^{\star}$PO (left) and
    $\mathrm{@}k$-GRPO (right) mitigate excessive sharpening for $k>1$.
    }
    \label{fig:code_lcb_maxk}
    \vspace{-0.6\baselineskip}
\end{wrapfigure}

\paragraph{Controlling sharpening by adjusting $k$.}
We observe the same qualitative effect as in the binary-reward
experiments. Figure~\ref{fig:code_lcb_maxk} compares the initial
checkpoint with win-rate-based $\mathrm{@k}$-$A^{\star}$PO and $\mathrm{@k}$-GRPO trained with 
$k\in\{1,2,4\}$.
The $k=1$ configuration typically provides the largest improvement for $\mathrm{max@}1$, but its advantage contracts as the evaluation budget increases and the starting checkpoint eventually catches up.
Training with $k>1$ retains (slightly) less of the $\mathrm{max@1}$ gain achieved by $k=1$ training, but overtakes it at larger evaluation budgets, maintaining a greater advantage over the starting checkpoint. Thus, our explicit win-rate-based reward transformation provides analogous control over sharpening for continuous rewards as our binary-reward adaptations.\looseness-1


\section{Conclusion}

We derive closed-form optimal policies for KL-regularized $\mathrm{pass@k}$ and $\mathrm{max@k}$ optimization, proving that $\mathrm{@k}$ optimization redistributes pressure according to task difficulty. For binary rewards, this pressure can be expressed equivalently through prompt-adaptive reward scaling or regularization strength. For continuous rewards, it appears as an adaptive reward transformation or a completion-dependent regularization strength, with a fully explicit construction for win-rate-based rewards. 
Leveraging these insights, we introduce \textsc{KAPPO}, a framework that adapts existing $\mathrm{@}1$ policy-optimization methods to $\mathrm{@k}$ settings without requiring online multi-completion rollouts. In particular, our $\mathrm{@k}$-$A^{\star}$PO instantiation requires only one new completion per prompt, yet achieves performance comparable to a SoTA multi-completion baseline. Validated across mathematical reasoning and code generation, $\mathrm{@k}$-$A^{\star}$PO, $\mathrm{@k}$-DPO, and $\mathrm{@k}$-GRPO successfully control sharpening, confirming that diverse reasoning can be effectively encouraged via prompt-adaptive optimization pressure.\looseness-1

\paragraph{Limitations and future work.}
Our reference calibration relies on a \emph{static} bank of $M$ completions for every prompt instead of, e.g., optimizing the budget based on the noise transfer of the equations in Propositions~\ref{prop:passk_fixed_point} and \ref{prop:winrate_maxk_optimal_policy}, resulting in noisier estimates for $p^*$ and $S_x^\mathrm{q}$. Future work may develop better budget allocations. Second, our continuous-reward experiments optimize the win-rate surrogate $S_x^{\mathrm{q}}$ rather than raw-reward $\mathrm{max@k}$, leaving empirical evaluation of numerical $F_{\pi^*}$ reconstruction (see Appendix~\ref{app:maxk_reconstruction}) to future work. Lastly, our win-rate solution assumes a continuous reference reward distribution; its application to discrete rewards with ties is an approximation.\looseness-1

\section*{Acknowledgements}

This work was supported as part of the Swiss AI Initiative by a grant from the Swiss National Supercomputing Centre (CSCS) under project ID infra01 on Alps. It was partially funded by an unrestricted gift from Coefficient Giving and by the Swiss National Science Foundation under grant number 212111. We are grateful to Robert West for his supervision and support.

\subsection*{AI use statement}

We used generative AI tools to check proofs, improve their
exposition, edit the manuscript, and assist with literature search
and reference checking. AI also supported the implementation of
methods, training infrastructure, evaluation, and plotting, following
author-specified designs and guidance. For data processing, AI assisted
with deduplication and analysis of dataset diversity, task difficulty,
and metadata.
The central research ideas and essential proof steps were developed
by the authors. All experimental decisions, including experimental design,
hyperparameters, comparisons, and all interpretations of results
were determined solely by the authors. The authors reviewed the AI-assisted
code, verified mathematical claims and proofs, checked cited sources
and their relevance, and checked figures against experimental outputs.
We take responsibility for the final content of this work, including
all AI-assisted text, claims, and artifacts.

\subsection*{Ethics statement}
We improve understanding of pass$\mathrm{@k}$ and max$\mathrm{@k}$ optimization and introduce a method to adapt existing reinforcement learning methods from $\mathrm{@1}$  to $\mathrm{@k}$. As a result, post-training of frontier models may be accelerated, which can result in all kinds of secondary effects, both economic and societal. Our experiments use only publicly available models and datasets in accordance with their licenses.

\subsection*{Reproducibility Statement}
All considered datasets and models are publicly available. Instead of proposing a novel RL optimization method with many degrees of freedom to be specified, we identify a minimalist intervention to turn any standard $\mathrm{@1}$ RL method into an $\mathrm{@k}$ method. We provide pseudocode for reference calibration with binary and continuous win-rate rewards in Appendix~\ref{app:calibration_algorithms}. As a result, all of our experiments can be easily reproduced. For completeness, we describe the exact experimental details in Appendix~\ref{app:experimental_details}. Our entire code base is available in the supplementary materials.

\bibliography{iclr2027_conference}
\bibliographystyle{iclr2027_conference}

\clearpage
\appendix
\addtocontents{toc}{\protect\setcounter{tocdepth}{2}}
\section*{Appendix}
\tableofcontents
\clearpage

\section{A Variational Optimality Lemma}
\label{app:variational_lemma}

We first establish a general result used in the derivations of both
optimal policies. Since all objectives considered in this work
decompose over prompts, we fix a prompt $x$ throughout this section.
Let
\begin{equation}
\mathcal{S}_x
=
\left\{
y:\pi_{\mathrm{ref}}(y\mid x)>0
\right\}
\end{equation}
denote the support of the reference policy. Any policy assigning
positive probability outside $\mathcal{S}_x$ has infinite KL
divergence, so it is sufficient to optimize over distributions
supported on $\mathcal{S}_x$.

Consider an objective of the form
\begin{equation}\label{eq:app_general_objective}
\mathcal{J}_x(\pi)
=
U_x(\pi)
-
\beta\,\mathrm{KL}
\left(
\pi(\cdot\mid x)
\Vert
\pi_{\mathrm{ref}}(\cdot\mid x)
\right),
\end{equation}
where $U_x$ is a concave reward functional and $\beta>0$.

For a feasible policy $\pi$, consider a perturbation
\begin{equation}\label{eq:app_policy_perturbation}
\pi_\varepsilon(y\mid x)
=
\pi(y\mid x)+\varepsilon h(y\mid x),
\end{equation}
where
\begin{equation}
\sum_{y\in\mathcal{S}_x}
\left|h(y\mid x)\right|
<\infty,
\qquad
\sum_{y\in\mathcal{S}_x}
h(y\mid x)
=
0,
\end{equation}
and there exists $\varepsilon_0>0$ such that
$\pi_\varepsilon(y\mid x)\geq0$ for every
$\varepsilon\in[0,\varepsilon_0]$ and
$y\in\mathcal{S}_x$. The first condition ensures that the perturbation
has finite total variation, the second preserves normalization, and
the third ensures that the perturbed policy remains feasible in the
direction under consideration.

We assume that the corresponding one-sided directional derivative of
$U_x$ exists and can be written as
\begin{equation}\label{eq:app_reward_directional_derivative}
\left.
\frac{d}{d\varepsilon}
U_x(\pi+\varepsilon h)
\right|_{\varepsilon=0^+}
=
\sum_{y\in\mathcal{S}_x}
h(y\mid x)A_\pi(y,x)
\end{equation}
for some score $A_\pi(y,x)$, with all displayed sums finite. The
specific score corresponding to each $\mathrm{@k}$ objective is
derived in the following sections.

We additionally assume that an optimal policy exists and has full
support on $\mathcal{S}_x$. Under the regularity conditions used below,
the reward change produced by transferring probability
$\varepsilon$ between two completions is $O(\varepsilon)$. Suppose,
however, that $\pi(y\mid x)=0$ for some
$y\in\mathcal{S}_x$ and transfer probability $\varepsilon$ to $y$.
The resulting change in the negative KL term contains
\begin{equation}
-\beta\varepsilon\log\varepsilon
=
\beta\varepsilon\log\frac{1}{\varepsilon}.
\end{equation}
Although this quantity tends to zero as
$\varepsilon\rightarrow0$, its ratio to $\varepsilon$ diverges.
Therefore, for sufficiently small $\varepsilon$, the KL gain dominates
the $O(\varepsilon)$ reward change, and a policy assigning zero
probability to part of $\mathcal{S}_x$ cannot be optimal.

\begin{lemma}[Variational form of a KL-regularized optimum]
\label{lem:variational_optimum}
Under the assumptions above, any maximizer $\pi^*$ of
Eq.~\ref{eq:app_general_objective} satisfies
\begin{equation}\label{eq:app_general_optimal_policy}
\pi^*(y\mid x)
=
\frac{1}{Z(x)}
\pi_{\mathrm{ref}}(y\mid x)
\exp\left(
\frac{A_{\pi^*}(y,x)}{\beta}
\right),
\end{equation}
where
\begin{equation}\label{eq:app_general_partition}
Z(x)
=
\sum_{y\in\mathcal{S}_x}
\pi_{\mathrm{ref}}(y\mid x)
\exp\left(
\frac{A_{\pi^*}(y,x)}{\beta}
\right).
\end{equation}
Conversely, any feasible policy satisfying
Eq.~\ref{eq:app_general_optimal_policy} is the unique maximizer of
Eq.~\ref{eq:app_general_objective}.
\end{lemma}

\begin{proof}
For compactness, define
\begin{equation}\label{eq:app_stationarity_score}
B_{\pi^*}(y,x)
=
A_{\pi^*}(y,x)
-
\beta
\left(
\log
\frac{\pi^*(y\mid x)}
     {\pi_{\mathrm{ref}}(y\mid x)}
+1
\right).
\end{equation}

Consider any perturbation $h$ such that
$\pi^*+\varepsilon h$ remains feasible for both positive and negative
sufficiently small $\varepsilon$. Applying
Eq.~\ref{eq:app_reward_directional_derivative} to $h$ and $-h$ shows
that the corresponding two-sided derivative exists. Since
$\varepsilon=0$ is a local maximum of
$\mathcal{J}_x(\pi^*+\varepsilon h)$, this derivative must vanish:
\begin{equation}\label{eq:app_stationarity_direction}
\left.
\frac{d}{d\varepsilon}
\mathcal{J}_x(\pi^*+\varepsilon h)
\right|_{\varepsilon=0}
=
0.
\end{equation}

The directional derivative of the KL term is
\begin{equation}\label{eq:app_kl_directional_derivative}
\left.
\frac{d}{d\varepsilon}
\mathrm{KL}
\left(
\pi^*+\varepsilon h
\Vert
\pi_{\mathrm{ref}}
\right)
\right|_{\varepsilon=0}
=
\sum_{y\in\mathcal{S}_x}
h(y\mid x)
\left(
\log
\frac{\pi^*(y\mid x)}
     {\pi_{\mathrm{ref}}(y\mid x)}
+1
\right).
\end{equation}
Together with
Eq.~\ref{eq:app_reward_directional_derivative}, this gives
\begin{equation}\label{eq:app_zero_directional_derivative}
\sum_{y\in\mathcal{S}_x}
h(y\mid x)B_{\pi^*}(y,x)
=
0.
\end{equation}

Now consider any two completions $y,z\in\mathcal{S}_x$ and choose
\begin{equation}
h
=
\mathbf{e}_y-\mathbf{e}_z,
\end{equation}
where $\mathbf{e}_y$ denotes a unit point mass at $y$. Since $\pi^*$
has full support, this perturbation remains feasible for both signs of
sufficiently small $\varepsilon$. Substituting it into
Eq.~\ref{eq:app_zero_directional_derivative} yields
\begin{equation}
B_{\pi^*}(y,x)-B_{\pi^*}(z,x)=0.
\end{equation}
Because $y$ and $z$ were arbitrary, $B_{\pi^*}(y,x)$ is constant
with respect to $y$ on $\mathcal{S}_x$. Hence, there exists a
prompt-dependent scalar $c(x)$ such that
\begin{equation}\label{eq:app_constant_stationarity}
A_{\pi^*}(y,x)
-
\beta
\left(
\log
\frac{\pi^*(y\mid x)}
     {\pi_{\mathrm{ref}}(y\mid x)}
+1
\right)
=
c(x)
\end{equation}
for every $y\in\mathcal{S}_x$.

Rearranging Eq.~\ref{eq:app_constant_stationarity} gives
\begin{equation}
\pi^*(y\mid x)
=
\pi_{\mathrm{ref}}(y\mid x)
\exp\left(
\frac{A_{\pi^*}(y,x)}{\beta}
-\frac{c(x)}{\beta}-1
\right).
\end{equation}
The final two terms are independent of $y$ and can therefore be
absorbed into the normalization constant. Imposing
$\sum_y\pi^*(y\mid x)=1$ gives
Eqs.~\ref{eq:app_general_optimal_policy}
and~\ref{eq:app_general_partition}.

It remains to show that this stationary policy is the unique global
maximizer. Let $\pi$ be any other feasible policy and define
\begin{equation}
h(y\mid x)
=
\pi(y\mid x)-\pi^*(y\mid x).
\end{equation}
This perturbation satisfies
\begin{equation}
\sum_y |h(y\mid x)|\leq2,
\qquad
\sum_y h(y\mid x)=0,
\end{equation}
and
\begin{equation}
\pi^*+\varepsilon h
=
(1-\varepsilon)\pi^*+\varepsilon\pi
\end{equation}
is a valid policy for every $\varepsilon\in[0,1]$.

Define
\begin{equation}
g(\varepsilon)
=
U_x(\pi^*+\varepsilon h).
\end{equation}
Because $U_x$ is concave, $g$ is concave on $[0,1]$. A concave
function lies below its supporting line at the origin, and hence
\begin{equation}
g(1)-g(0)
\leq
g'(0^+).
\end{equation}
Using Eq.~\ref{eq:app_reward_directional_derivative}, this gives
\begin{equation}\label{eq:app_reward_concavity_bound}
U_x(\pi)-U_x(\pi^*)
\leq
\sum_{y\in\mathcal{S}_x}
\left(
\pi(y\mid x)-\pi^*(y\mid x)
\right)
A_{\pi^*}(y,x).
\end{equation}

The KL difference admits the exact decomposition
\begin{align}
&
\mathrm{KL}
\left(
\pi\Vert\pi_{\mathrm{ref}}
\right)
-
\mathrm{KL}
\left(
\pi^*\Vert\pi_{\mathrm{ref}}
\right)
\nonumber\\
&\quad=
\sum_{y\in\mathcal{S}_x}
\left(
\pi(y\mid x)-\pi^*(y\mid x)
\right)
\left(
\log
\frac{\pi^*(y\mid x)}
     {\pi_{\mathrm{ref}}(y\mid x)}
+1
\right)
+
\mathrm{KL}
\left(
\pi\Vert\pi^*
\right).
\label{eq:app_kl_three_point_identity}
\end{align}
To verify this identity, expand
\begin{equation}
\log
\frac{\pi(y\mid x)}
     {\pi_{\mathrm{ref}}(y\mid x)}
=
\log
\frac{\pi(y\mid x)}
     {\pi^*(y\mid x)}
+
\log
\frac{\pi^*(y\mid x)}
     {\pi_{\mathrm{ref}}(y\mid x)}
\end{equation}
inside the first KL term and use
$\sum_y(\pi(y\mid x)-\pi^*(y\mid x))=0$.

Combining Eqs.~\ref{eq:app_reward_concavity_bound}
and~\ref{eq:app_kl_three_point_identity} yields
\begin{align}
\mathcal{J}_x(\pi)-\mathcal{J}_x(\pi^*)
\leq
&
\sum_{y\in\mathcal{S}_x}
\left(
\pi(y\mid x)-\pi^*(y\mid x)
\right)
B_{\pi^*}(y,x)
\nonumber\\
&-
\beta\,
\mathrm{KL}
\left(
\pi\Vert\pi^*
\right).
\end{align}
By Eq.~\ref{eq:app_constant_stationarity},
$B_{\pi^*}(y,x)=c(x)$ for every $y$. Since both policies have total
mass one,
\begin{equation}
\sum_{y\in\mathcal{S}_x}
\left(
\pi(y\mid x)-\pi^*(y\mid x)
\right)
B_{\pi^*}(y,x)
=
c(x)
\sum_{y\in\mathcal{S}_x}
\left(
\pi(y\mid x)-\pi^*(y\mid x)
\right)
=
0.
\end{equation}
Consequently,
\begin{equation}
\mathcal{J}_x(\pi)-\mathcal{J}_x(\pi^*)
\leq
-\beta\,
\mathrm{KL}
\left(
\pi\Vert\pi^*
\right)
\leq 0.
\end{equation}
Equality holds only if $\pi=\pi^*$, proving that $\pi^*$ is the unique
global maximizer.
\end{proof}

The score $A_\pi(y,x)$ is defined only up to an additive term that
depends on $x$ but not on $y$. Such a term vanishes in
Eq.~\ref{eq:app_reward_directional_derivative} because every permitted
perturbation has zero total mass, and it is absorbed into $Z(x)$ in
Eq.~\ref{eq:app_general_optimal_policy}. The following sections derive
the relevant score for the $\mathrm{pass@k}$ and $\mathrm{max@k}$
reward functionals.

\section{Binary-Rewards Derivation}
\label{app:passk_derivation}

We now apply Lemma~\ref{lem:variational_optimum} to the $\mathrm{pass@k}$ objective. As before, the objective decomposes over prompts, so we fix a prompt $x$. For any policy $\pi$, define its success probability
\begin{equation}
p_\pi(x)
=
\mathbb{P}_{y\sim\pi(\cdot\mid x)}[R(y,x)=1]
=
\sum_y \pi(y\mid x)R(y,x).
\end{equation}
For binary rewards, the probability that at least one of $k$ independent completions succeeds is
\begin{equation}
U_x(\pi)
=
1-\left(1-p_\pi(x)\right)^k.
\end{equation}
The prompt-wise KL-regularized objective is therefore
\begin{equation}\label{eq:app_passk_objective}
\mathcal{J}_{k,x}(\pi)
=
1-\left(1-p_\pi(x)\right)^k
-
\beta\,\mathrm{KL}
\left(
\pi(\cdot\mid x)
\Vert
\pi_{\mathrm{ref}}(\cdot\mid x)
\right).
\end{equation}

\subsection{Optimal Policy}

\noindent\textbf{Proposition~\ref{prop:passk_optimal_policy} (restated).}
For $R(y,x)\in\{0,1\}$, the unique maximizer of $\mathcal{J}_{k,x}$ satisfies
\[
\pi^*(y\mid x)
=
\frac{1}{Z(x)}
\pi_{\mathrm{ref}}(y\mid x)
\exp\left(\frac{R(y,x)}{\beta_{\mathrm{eff}}(x)}\right),
\]
where
\[
\beta_{\mathrm{eff}}(x)
=
\frac{\beta}{k(1-p^*(x))^{k-1}},
\qquad
Z(x)
=
1-p_{\mathrm{ref}}(x)
+
p_{\mathrm{ref}}(x)e^{1/\beta_{\mathrm{eff}}(x)},
\]
with $p_{\mathrm{ref}}(x)=p_{\pi_{\mathrm{ref}}}(x)$ and $p^*(x)=p_{\pi^*}(x)$.

\begin{proof}
We first consider $0<p_{\mathrm{ref}}(x)<1$; the boundary cases are treated below. The mapping $p\mapsto1-(1-p)^k$ is concave on $[0,1]$ for $k\geq1$: it is affine for $k=1$, while for $k\geq2$,
\begin{equation}
\frac{d^2}{dp^2}\left[1-(1-p)^k\right]
=
-k(k-1)(1-p)^{k-2}
\leq0.
\end{equation}
Since $p_\pi(x)$ is linear in $\pi$, the reward functional $U_x(\pi)$ is concave. We can therefore apply Lemma~\ref{lem:variational_optimum} under its stated regularity assumptions.

Consider a perturbation $\pi_\varepsilon=\pi+\varepsilon h$ satisfying the conditions of Appendix~\ref{app:variational_lemma}. Its success probability is
\begin{equation}
p_{\pi_\varepsilon}(x)
=
p_\pi(x)
+
\varepsilon\sum_y h(y\mid x)R(y,x).
\end{equation}
Consequently, the directional derivative of the reward functional is
\begin{align}
\left.
\frac{d}{d\varepsilon}
U_x(\pi+\varepsilon h)
\right|_{\varepsilon=0^+}
&=
k\left(1-p_\pi(x)\right)^{k-1}
\sum_y h(y\mid x)R(y,x).
\end{align}
Thus, the score in Eq.~\ref{eq:app_reward_directional_derivative} is
\begin{equation}\label{eq:app_passk_score}
A_\pi(y,x)
=
k\left(1-p_\pi(x)\right)^{k-1}R(y,x).
\end{equation}

Applying Lemma~\ref{lem:variational_optimum} gives
\begin{equation}\label{eq:app_passk_implicit_policy}
\pi^*(y\mid x)
=
\frac{1}{Z(x)}
\pi_{\mathrm{ref}}(y\mid x)
\exp\left(
\frac{k}{\beta}
\left(1-p^*(x)\right)^{k-1}R(y,x)
\right),
\end{equation}
where
\begin{equation}
p^*(x)=p_{\pi^*}(x).
\end{equation}

Define
\begin{equation}
p_{\mathrm{ref}}(x)
=
\mathbb{P}_{y\sim\pi_{\mathrm{ref}}(\cdot\mid x)}
[R(y,x)=1].
\end{equation}
Since the reward is binary, the reference policy assigns total mass $1-p_{\mathrm{ref}}(x)$ to unsuccessful completions and total mass $p_{\mathrm{ref}}(x)$ to successful completions. Normalizing Eq.~\ref{eq:app_passk_implicit_policy} therefore gives
\begin{equation}\label{eq:app_passk_partition}
Z(x)
=
1-p_{\mathrm{ref}}(x)
+
p_{\mathrm{ref}}(x)
\exp\left(
\frac{k}{\beta}\left(1-p^*(x)\right)^{k-1}
\right).
\end{equation}

Introducing
\begin{equation}\label{eq:app_passk_effective_beta}
\beta_{\mathrm{eff}}(x)
=
\frac{\beta}{k\left(1-p^*(x)\right)^{k-1}},
\end{equation}
we recover the form reported in Eq.~\ref{eq:passk_optimal_policy}:
\begin{equation}
\pi^*(y\mid x)
=
\frac{1}{Z(x)}
\pi_{\mathrm{ref}}(y\mid x)
\exp\left(
\frac{R(y,x)}{\beta_{\mathrm{eff}}(x)}
\right).
\qedhere
\end{equation}
\end{proof}

The solution remains implicit because its right-hand side depends on the success probability of the unknown optimal policy.

\subsection{Scalar Equation and Uniqueness}

It remains to determine $p^*(x)$.

\noindent\textbf{Proposition~\ref{prop:passk_fixed_point} (restated).}
For $0<p_{\mathrm{ref}}(x)<1$, the optimal success probability $p^*(x)$ is the unique solution in $(0,1)$ of
\[
k(1-p^*(x))^{k-1}
=
\beta\log
\frac{p^*(x)(1-p_{\mathrm{ref}}(x))}
{p_{\mathrm{ref}}(x)(1-p^*(x))}.
\]

\begin{proof}
Summing Eq.~\ref{eq:app_passk_implicit_policy} over successful completions gives
\begin{equation}\label{eq:app_passk_success_relation}
p^*(x)
=
\frac{
p_{\mathrm{ref}}(x)
\exp\left(\frac{k}{\beta}(1-p^*(x))^{k-1}\right)
}{
1-p_{\mathrm{ref}}(x)
+
p_{\mathrm{ref}}(x)
\exp\left(\frac{k}{\beta}(1-p^*(x))^{k-1}\right)
}.
\end{equation}
Taking the odds ratio and then the logarithm yields
\begin{equation}\label{eq:app_passk_fixed_point}
k\left(1-p^*(x)\right)^{k-1}
=
\beta\log
\frac{
p^*(x)\left(1-p_{\mathrm{ref}}(x)\right)
}{
p_{\mathrm{ref}}(x)\left(1-p^*(x)\right)
},
\end{equation}
which is Eq.~\ref{eq:passk_fixed_point} in the main text.

To establish uniqueness, fix $q=p_{\mathrm{ref}}(x)\in(0,1)$ and define, for $p\in(0,1)$,
\begin{equation}
\Phi_q(p)
=
\beta\log\frac{p(1-q)}{q(1-p)}
-
k(1-p)^{k-1}.
\end{equation}
Equation~\ref{eq:app_passk_fixed_point} is equivalent to $\Phi_q(p)=0$. The function satisfies
\begin{equation}
\lim_{p\rightarrow0^+}\Phi_q(p)=-\infty,
\qquad
\lim_{p\rightarrow1^-}\Phi_q(p)=+\infty,
\end{equation}
and
\begin{equation}
\Phi_q'(p)
=
\frac{\beta}{p(1-p)}
+
k(k-1)(1-p)^{k-2}
>
0.
\end{equation}
Hence, $\Phi_q$ is continuous and strictly increasing from $-\infty$ to $+\infty$, so it has exactly one root in $(0,1)$. Moreover,
\begin{equation}
\Phi_q(q)
=
-k(1-q)^{k-1}
<0,
\end{equation}
which implies that the unique solution satisfies
\begin{equation}
p^*(x)>p_{\mathrm{ref}}(x).
\qedhere
\end{equation}
\end{proof}

For $k=1$, Eq.~\ref{eq:app_passk_fixed_point} reduces to
\begin{equation}
1
=
\beta\log
\frac{
p^*(x)\left(1-p_{\mathrm{ref}}(x)\right)
}{
p_{\mathrm{ref}}(x)\left(1-p^*(x)\right)
},
\end{equation}
whose solution is
\begin{equation}
p^*(x)
=
\frac{
p_{\mathrm{ref}}(x)e^{1/\beta}
}{
1-p_{\mathrm{ref}}(x)
+
p_{\mathrm{ref}}(x)e^{1/\beta}
}.
\end{equation}
In this case, $\beta_{\mathrm{eff}}(x)=\beta$, and Eq.~\ref{eq:app_passk_implicit_policy} reduces to the standard KL-regularized single-reward optimal policy.

\subsection{Boundary Cases}

The scalar equation above assumes $0<p_{\mathrm{ref}}(x)<1$. The two boundary cases follow directly from the support constraint imposed by the KL term.

\noindent\textbf{Boundary cases.}
If $p_{\mathrm{ref}}(x)\in\{0,1\}$, the unique maximizer of $\mathcal{J}_{k,x}$ is the reference policy itself:
\[
\pi^*(\cdot\mid x)=\pi_{\mathrm{ref}}(\cdot\mid x),
\qquad
p^*(x)=p_{\mathrm{ref}}(x).
\]

\begin{proof}
If $p_{\mathrm{ref}}(x)=0$, the reference policy assigns no probability to successful completions. Every policy with finite KL divergence is therefore also supported only on unsuccessful completions, so
\begin{equation}
p^*(x)=0,
\qquad
\pi^*(\cdot\mid x)=\pi_{\mathrm{ref}}(\cdot\mid x).
\end{equation}
The reward term is identically zero over the feasible set, and the reference policy uniquely minimizes the KL term.

If $p_{\mathrm{ref}}(x)=1$, every completion in the support of the reference policy is successful. Hence, every finite-KL policy has success probability one and the $\mathrm{pass@k}$ reward is constant:
\begin{equation}
p^*(x)=1,
\qquad
\pi^*(\cdot\mid x)=\pi_{\mathrm{ref}}(\cdot\mid x).
\end{equation}
The reference policy again uniquely minimizes the KL term.
\end{proof}

For $k>1$, the case $p_{\mathrm{ref}}(x)=1$ corresponds to $\beta_{\mathrm{eff}}(x)=+\infty$: once the task is solved with probability one, the $\mathrm{pass@k}$ reward provides no further optimization pressure. For $k=1$, the effective regularization remains $\beta$.

\subsection{Objective-Level Equivalences}

We finally prove the two objective-level interpretations stated in the main text.

\noindent\textbf{Corollary~\ref{cor:passk_adaptive_regularization} (restated).}
Replacing the constant KL coefficient $\beta$ in the ordinary single-reward objective with $\beta_{\mathrm{eff}}(x)$ from Eq.~\ref{eq:app_passk_effective_beta} recovers the $\mathrm{pass@k}$ optimal policy, with $\beta_{\mathrm{eff}}(x)=+\infty$ interpreted in the limiting sense.

\begin{proof}
For each prompt $x$ with finite $\beta_{\mathrm{eff}}(x)$, consider the ordinary single-reward objective with this prompt-dependent KL coefficient. Its unique optimal policy is
\begin{equation}
\pi_{\mathrm{reg}}^*(y\mid x)
=
\frac{1}{Z_{\mathrm{reg}}(x)}
\pi_{\mathrm{ref}}(y\mid x)
\exp\left(
\frac{R(y,x)}{\beta_{\mathrm{eff}}(x)}
\right).
\end{equation}
This is exactly the $\mathrm{pass@k}$ optimal policy derived in Eq.~\ref{eq:app_passk_implicit_policy}. Since the objective decomposes over prompts and $\beta_{\mathrm{eff}}(x)$ is constant with respect to $y$ for each fixed $x$, the two objectives have the same optimal policy. When $\beta_{\mathrm{eff}}(x)=+\infty$, the limiting policy is $\pi_{\mathrm{ref}}$, agreeing with the boundary case above.
\end{proof}

\noindent\textbf{Corollary~\ref{cor:passk_adaptive_reward} (restated).}
The ordinary single-reward objective with constant KL coefficient $\beta$ and transformed reward
\[
\widetilde R_k(y,x)
=
\frac{\beta}{\beta_{\mathrm{eff}}(x)}R(y,x)
=
k(1-p^*(x))^{k-1}R(y,x)
\]
has the same unique optimal policy as the $\mathrm{pass@k}$ objective.

\begin{proof}
Retain the constant KL coefficient $\beta$ and replace the original binary reward with
\begin{equation}
\widetilde R_k(y,x)
=
\frac{\beta}{\beta_{\mathrm{eff}}(x)}R(y,x)
=
k\left(1-p^*(x)\right)^{k-1}R(y,x).
\end{equation}
The corresponding single-reward optimal policy is
\begin{align}
\pi_{\mathrm{rew}}^*(y\mid x)
&=
\frac{1}{Z_{\mathrm{rew}}(x)}
\pi_{\mathrm{ref}}(y\mid x)
\exp\left(
\frac{\widetilde R_k(y,x)}{\beta}
\right)\\
&=
\frac{1}{Z_{\mathrm{rew}}(x)}
\pi_{\mathrm{ref}}(y\mid x)
\exp\left(
\frac{R(y,x)}{\beta_{\mathrm{eff}}(x)}
\right).
\end{align}
This again coincides exactly with the $\mathrm{pass@k}$ optimal policy.
\end{proof}

\section{Continuous-Rewards Derivation}
\label{app:maxk_derivation}

We now derive the optimal policy for the $\mathrm{max@k}$ objective with real-valued rewards. As before, the objective decomposes over prompts, so we fix a prompt $x$. For the derivation, it is sufficient to assume that
\begin{equation}
R(y,x)\in[a_x,b_x]
\end{equation}
for every $y$ in the support of the reference policy, where the endpoints may depend on $x$. A common reward range across prompts is not required for the optimal-policy derivation; it becomes relevant only for the cross-prompt interpretation discussed in Appendix~\ref{app:maxk_reward_scale}.

Boundedness guarantees that the relevant expectations and directional derivatives are finite and permits differentiation under the expectation. The result extends to unbounded rewards whenever the corresponding expectations, integrals, and directional derivatives are finite, the required interchange of differentiation and expectation is valid, and the optimum is attained.

No continuity assumption on the reward distribution is required in this section. In particular, the reward CDF may contain atoms.

For a policy $\pi$, define
\begin{equation}
U_{k,x}(\pi)
=
\mathbb{E}_{y_1,\ldots,y_k\iid\pi(\cdot\mid x)}
\left[
\max_{i=1,\ldots,k}R(y_i,x)
\right].
\end{equation}
The prompt-wise KL-regularized objective is
\begin{equation}\label{eq:app_maxk_objective}
\mathcal{J}_{k,x}(\pi)
=
U_{k,x}(\pi)
-
\beta\,\mathrm{KL}
\left(
\pi(\cdot\mid x)
\Vert
\pi_{\mathrm{ref}}(\cdot\mid x)
\right).
\end{equation}

\subsection{Optimal Policy}

\noindent\textbf{Proposition~\ref{prop:maxk_adaptive_reward} (restated).}
For $k>1$ and $\beta>0$, under the assumptions above, the unique maximizer of $\mathcal{J}_{k,x}$ satisfies
\[
\pi^*(y\mid x)
=
\frac{1}{Z(x)}
\pi_{\mathrm{ref}}(y\mid x)
\exp\left(
\frac{k}{\beta}
\int_{-\infty}^{R(y,x)}
F_{\pi^*}(r\mid x)^{k-1}\,dr
\right),
\]
where $F_{\pi^*}(r\mid x)=\mathbb{P}_{y\sim\pi^*(\cdot\mid x)}[R(y,x)\leq r]$ is the optimal reward CDF and $Z(x)$ normalizes the policy.

\begin{proof}
\textit{Step 1: Concavity and existence.}
Let
\begin{equation}
F_\pi(r\mid x)
=
\mathbb{P}_{y\sim\pi(\cdot\mid x)}[R(y,x)\leq r]
\end{equation}
denote the reward CDF under $\pi$. Since the maximum of $k$ independent rewards is bounded below by $a_x$, the tail-integral identity gives
\begin{equation}\label{eq:app_maxk_cdf_objective}
U_{k,x}(\pi)
=
a_x
+
\int_{a_x}^{b_x}
\left[1-F_\pi(r\mid x)^k\right]dr.
\end{equation}
For each fixed $r$, $F_\pi(r\mid x)$ is linear in $\pi$, while the mapping
\begin{equation}
u\longmapsto1-u^k
\end{equation}
is concave on $[0,1]$. Integration preserves concavity, so $U_{k,x}(\pi)$ is a concave functional of $\pi$. Combined with the strict concavity of the negative KL term, this implies that Eq.~\ref{eq:app_maxk_objective} has at most one maximizer.

For a finite completion space, existence follows from compactness of the probability simplex. For a countably infinite space, boundedness of the reward allows a maximizing sequence to be chosen with uniformly bounded KL divergence from $\pi_{\mathrm{ref}}$. The corresponding KL sublevel sets are tight, so this sequence has a subsequence converging in total variation. The bounded reward functional is continuous under this convergence, while KL is lower semicontinuous, ensuring that the supremum is attained. Bounded rewards also satisfy the full-support condition discussed in Appendix~\ref{app:variational_lemma}.

\medskip
\noindent\textit{Step 2: Directional derivative.}
Consider a perturbation
\begin{equation}
\pi_\varepsilon=\pi+\varepsilon h
\end{equation}
satisfying the conditions of Appendix~\ref{app:variational_lemma}. Expanding the product measure in $U_{k,x}(\pi_\varepsilon)$ and differentiating at $\varepsilon=0$ gives one term for each of the $k$ sampled completions. By symmetry,
\begin{align}
\left.
\frac{d}{d\varepsilon}
U_{k,x}(\pi+\varepsilon h)
\right|_{\varepsilon=0^+}
&=
k\sum_y h(y\mid x)
\mathbb{E}_{y_2,\ldots,y_k\iid\pi(\cdot\mid x)}
\left[
\max\left\{
R(y,x),R(y_2,x),\ldots,R(y_k,x)
\right\}
\right].
\label{eq:app_maxk_directional_derivative}
\end{align}
Thus, one valid score in Eq.~\ref{eq:app_reward_directional_derivative} is
\begin{equation}\label{eq:app_maxk_initial_score}
\Gamma_\pi(y,x)
=
k\,\mathbb{E}_{y_2,\ldots,y_k\iid\pi(\cdot\mid x)}
\left[
\max\left\{
R(y,x),R(y_2,x),\ldots,R(y_k,x)
\right\}
\right].
\end{equation}

To isolate the part that depends on $y$, let
\begin{equation}
M_{k-1}
=
\max_{i=2,\ldots,k}R(y_i,x).
\end{equation}
Using
\begin{equation}
\max\{a,b\}=b+(a-b)_+,
\qquad
(u)_+=\max\{u,0\},
\end{equation}
we obtain
\begin{align}
\Gamma_\pi(y,x)
&=
k\,\mathbb{E}_{\pi}[M_{k-1}]
+
k\,\mathbb{E}_{\pi}
\left[
\left(R(y,x)-M_{k-1}\right)_+
\right].
\label{eq:app_maxk_score_decomposition}
\end{align}
The first term depends on $\pi$ and $x$ but not on the candidate completion $y$. It can therefore be omitted from the score by the additive-constant invariance established after Lemma~\ref{lem:variational_optimum}.

For any real number $a$ and random variable $M$,
\begin{equation}
(a-M)_+
=
\int_{-\infty}^{a}\mathbb{I}[M\leq r]\,dr.
\end{equation}
Taking expectations and applying Tonelli's theorem gives
\begin{equation}\label{eq:app_positive_part_identity}
\mathbb{E}[(a-M)_+]
=
\int_{-\infty}^{a}\mathbb{P}[M\leq r]\,dr.
\end{equation}
This identity holds for arbitrary reward distributions and does not require their CDFs to be continuous.

Because $M_{k-1}$ is the maximum of $k-1$ independent rewards sampled from $\pi$,
\begin{equation}
\mathbb{P}[M_{k-1}\leq r]
=
F_\pi(r\mid x)^{k-1}.
\end{equation}
Substituting this into Eq.~\ref{eq:app_positive_part_identity}, with $a=R(y,x)$, gives
\begin{equation}
\mathbb{E}_{\pi}
\left[
\left(R(y,x)-M_{k-1}\right)_+
\right]
=
\int_{-\infty}^{R(y,x)}
F_\pi(r\mid x)^{k-1}\,dr.
\end{equation}
We may therefore choose the completion-dependent score
\begin{equation}\label{eq:app_maxk_score}
A_\pi(y,x)
=
k\int_{-\infty}^{R(y,x)}
F_\pi(r\mid x)^{k-1}\,dr.
\end{equation}

\medskip
\noindent\textit{Step 3: Optimal policy.}
Applying Lemma~\ref{lem:variational_optimum} with the score in Eq.~\ref{eq:app_maxk_score} gives
\begin{equation}\label{eq:app_maxk_optimal_policy}
\pi^*(y\mid x)
=
\frac{1}{Z(x)}
\pi_{\mathrm{ref}}(y\mid x)
\exp\left(
\frac{k}{\beta}
\int_{-\infty}^{R(y,x)}
F_{\pi^*}(r\mid x)^{k-1}\,dr
\right),
\end{equation}
where
\begin{equation}
Z(x)
=
\sum_y
\pi_{\mathrm{ref}}(y\mid x)
\exp\left(
\frac{k}{\beta}
\int_{-\infty}^{R(y,x)}
F_{\pi^*}(r\mid x)^{k-1}\,dr
\right).
\end{equation}
This is Eq.~\ref{eq:maxk_optimal_policy_implicit} from the main text. By the converse statement of Lemma~\ref{lem:variational_optimum}, any feasible policy satisfying this characterization is the unique global maximizer.
\end{proof}

The characterization remains implicit because the CDF $F_{\pi^*}$ appearing on its right-hand side is itself induced by the unknown optimal policy. A numerical procedure for reconstructing $F_{\pi^*}$ from the reference reward distribution is provided in Appendix~\ref{app:maxk_reconstruction}.

\subsection{Objective-Level Equivalence}

\noindent\textbf{Corollary~\ref{cor:maxk_adaptive_reward} (restated).}
For $k>1$, define the fixed reward transformation
\[
S_x(r)
=
k\int_{-\infty}^{r}
F_{\pi^*}(u\mid x)^{k-1}\,du.
\]
The ordinary single-reward objective with constant KL coefficient $\beta$ and transformed reward $\widetilde R(y,x)=S_x(R(y,x))$ has the same unique optimal policy as the $\mathrm{max@k}$ objective.

\begin{proof}
Treating $S_x$ as a fixed reward transformation, the corresponding single-reward optimal policy is
\[
\pi_{\mathrm{rew}}^*(y\mid x)
=
\frac{1}{Z_{\mathrm{rew}}(x)}
\pi_{\mathrm{ref}}(y\mid x)
\exp\left(
\frac{S_x(R(y,x))}{\beta}
\right).
\]
Substituting the definition of $S_x$ gives exactly the exponent in Eq.~\ref{eq:app_maxk_optimal_policy}. Normalization therefore yields $Z_{\mathrm{rew}}(x)=Z(x)$ and $\pi_{\mathrm{rew}}^*=\pi^*$. Since both objectives decompose over prompts, the equivalence holds for the full objectives as well.
\end{proof}

\section{Numerical Reconstruction of the Raw-Reward Optimal CDF}
\label{app:maxk_reconstruction}

The optimal policy in
Eq.~\ref{eq:app_maxk_optimal_policy} depends on the unknown reward CDF
$F_{\pi^*}$. In this section, we show how this CDF can be reconstructed
numerically from the reference reward distribution. As in the previous
derivations, we fix a prompt $x$ and suppress its dependence where
doing so does not cause ambiguity.

Define the optimal-policy score
\begin{equation}\label{eq:app_maxk_reconstruction_score}
H(r)
=
k\int_{-\infty}^{r}
F_{\pi^*}(u\mid x)^{k-1}\,du
\end{equation}
and let
\begin{equation}
c(x)=\frac{1}{Z(x)}.
\end{equation}
Because the likelihood ratio in
Eq.~\ref{eq:app_maxk_optimal_policy} depends on a completion only
through its reward, the reward distributions under the optimal and
reference policies satisfy
\begin{equation}\label{eq:app_maxk_reward_distribution_relation}
dP_{\pi^*}(r\mid x)
=
c(x)
\exp\left(
\frac{H(r)}{\beta}
\right)
dP_{\mathrm{ref}}(r\mid x).
\end{equation}
Thus, once the scalar $c(x)$ is known, the optimal reward distribution
can be reconstructed progressively from lower to higher rewards. The
normalization condition determines the unique value of $c(x)$.

\subsection{Continuous Reference Reward Distribution}

Suppose first that the reference reward distribution has density
$f_{\mathrm{ref}}(r\mid x)$ on a bounded interval $[a_x,b_x]$. Let
$f_{\pi^*}(r\mid x)$ denote the corresponding optimal density.
Equation~\ref{eq:app_maxk_reward_distribution_relation} gives
\begin{equation}\label{eq:app_maxk_density_relation}
f_{\pi^*}(r\mid x)
=
c(x)
f_{\mathrm{ref}}(r\mid x)
\exp\left(
\frac{H(r)}{\beta}
\right).
\end{equation}
Since
\begin{equation}
\frac{d}{dr}F_{\pi^*}(r\mid x)
=
f_{\pi^*}(r\mid x),
\qquad
\frac{d}{dr}H(r)
=
kF_{\pi^*}(r\mid x)^{k-1},
\end{equation}
the unknown functions satisfy the coupled system
\begin{equation}\label{eq:app_maxk_reconstruction_ode}
\begin{aligned}
\frac{d}{dr}F_{\pi^*}(r\mid x)
&=
c(x)
f_{\mathrm{ref}}(r\mid x)
\exp\left(
\frac{H(r)}{\beta}
\right),\\
\frac{d}{dr}H(r)
&=
kF_{\pi^*}(r\mid x)^{k-1},
\end{aligned}
\end{equation}
with initial conditions
\begin{equation}
F_{\pi^*}(a_x\mid x)=0,
\qquad
H(a_x)=0.
\end{equation}
For a candidate value of $c(x)$, the system can be integrated forward
from $a_x$ to $b_x$. The correct value is the one satisfying the
terminal normalization condition
\begin{equation}\label{eq:app_maxk_reconstruction_boundary}
F_{\pi^*}(b_x\mid x)=1.
\end{equation}
This gives a one-dimensional shooting problem. The terminal value
$F_{\pi^*}(b_x\mid x)$ increases monotonically with $c(x)$, so the
normalizing value can be found using bisection or another scalar
root-finding method.

\subsection{Discrete Reconstruction}

In practice, the reference reward distribution is estimated from a
finite bank of reference-model completions and is therefore discrete.
Let
\begin{equation}
r_1<r_2<\cdots<r_m
\end{equation}
be its distinct observed reward values, and define
\begin{equation}
q_j
=
\mathbb{P}_{\pi_{\mathrm{ref}}}
\left[
R(y,x)=r_j
\right],
\qquad
j=1,\ldots,m.
\end{equation}
Let
\begin{equation}
p_j
=
\mathbb{P}_{\pi^*}
\left[
R(y,x)=r_j
\right],
\qquad
P_j
=
\sum_{\ell=1}^{j}p_\ell
\end{equation}
denote the optimal probability mass at reward $r_j$ and the
corresponding cumulative probability.

Define
\begin{equation}
H_j
=
H(r_j)
=
k\int_{-\infty}^{r_j}
F_{\pi^*}(u\mid x)^{k-1}\,du.
\end{equation}
Since the optimal CDF is constant between consecutive reward values,
\begin{equation}\label{eq:app_maxk_discrete_score_recursion}
H_1=0,
\qquad
H_j
=
H_{j-1}
+
k(r_j-r_{j-1})P_{j-1}^{k-1},
\quad j=2,\ldots,m.
\end{equation}
Grouping Eq.~\ref{eq:app_maxk_optimal_policy} by reward value gives
\begin{equation}\label{eq:app_maxk_discrete_mass}
p_j
=
c(x)q_j
\exp\left(
\frac{H_j}{\beta}
\right).
\end{equation}

Equations~\ref{eq:app_maxk_discrete_score_recursion}
and~\ref{eq:app_maxk_discrete_mass} yield a forward recursion for any
candidate $c=c(x)$:
\begin{equation}\label{eq:app_maxk_discrete_recursion}
\begin{aligned}
H_1(c)&=0,\\
p_1(c)&=cq_1,\\
P_1(c)&=p_1(c),\\
H_j(c)&=
H_{j-1}(c)
+
k(r_j-r_{j-1})P_{j-1}(c)^{k-1},\\
p_j(c)&=
cq_j\exp\left(\frac{H_j(c)}{\beta}\right),\\
P_j(c)&=
P_{j-1}(c)+p_j(c),
\qquad j=2,\ldots,m.
\end{aligned}
\end{equation}
The required normalization is
\begin{equation}\label{eq:app_maxk_discrete_normalization}
P_m(c)=1.
\end{equation}

Each quantity in the recursion is continuous and strictly increasing
in $c$. Moreover,
\begin{equation}
P_m(0)=0,
\end{equation}
while, for $q_1>0$,
\begin{equation}
P_m\left(\frac{1}{q_1}\right)\geq
p_1\left(\frac{1}{q_1}\right)=1,
\end{equation}
with strict inequality whenever $m>1$. Therefore,
Eq.~\ref{eq:app_maxk_discrete_normalization} has a unique solution in
\begin{equation}
0<c\leq\frac{1}{q_1},
\end{equation}
which can be found efficiently by bisection. Equivalently, one may
treat the optimal probability of the lowest reward,
$p_1=cq_1$, as the scalar search variable.

Once $c$ has been determined, the recursion recovers all optimal
masses $p_j$, cumulative probabilities $P_j$, and scores $H_j$. For
$r\in[r_j,r_{j+1})$, the reconstructed CDF and score are
\begin{equation}\label{eq:app_maxk_discrete_interpolation}
F_{\pi^*}(r\mid x)
=
P_j,
\qquad
H(r)
=
H_j
+
k(r-r_j)P_j^{k-1}.
\end{equation}
Thus, the score is linearly interpolated between adjacent reward
levels, with slope determined by the reconstructed optimal CDF.

\subsection{Empirical Approximation}

Given $M$ reference-model completions, the discrete reference masses
are estimated as
\begin{equation}
\widehat q_j
=
\frac{1}{M}
\sum_{i=1}^{M}
\mathbb{I}
\left[
R(y_i^{\mathrm{ref}},x)=r_j
\right].
\end{equation}
Substituting these empirical masses into
Eq.~\ref{eq:app_maxk_discrete_recursion} produces an approximation of
$F_{\pi^*}$ and the raw-reward transformation $H$. The reconstruction
must be performed once for each prompt and each pair $(k,\beta)$.

Unlike the binary-reward case, where the reference distribution is
summarized by the single scalar $p_{\mathrm{ref}}(x)$, the continuous
case requires estimating the complete reference reward distribution.
Errors in its estimated masses and reward locations propagate through
the forward recursion and the subsequent exponential reweighting.
This motivates the explicit reference-win-rate construction derived in
Appendix~\ref{app:winrate_maxk_derivation}.

\section{Reward-Scale Assumptions for Continuous Rewards}
\label{app:maxk_reward_scale}

The derivation in Appendix~\ref{app:maxk_derivation} requires only
prompt-wise bounded rewards, or suitable integrability conditions in
the unbounded case. The stronger shared-range assumption used in the
main text is needed only for interpreting the resulting adaptation
across prompts.

\paragraph{Shift invariance.}
Both objectives studied in this work are invariant, at the level of
their optimizer, to prompt-dependent constant reward shifts. For
\begin{equation}
R_c(y,x)=R(y,x)+c(x),
\end{equation}
we have
\begin{equation}
\mathbb{E}_{y\sim\pi(\cdot\mid x)}[R_c(y,x)]
=
\mathbb{E}_{y\sim\pi(\cdot\mid x)}[R(y,x)]+c(x)
\end{equation}
and
\begin{equation}
\mathbb{E}
\left[
\max_{i=1,\ldots,k}R_c(y_i,x)
\right]
=
\mathbb{E}
\left[
\max_{i=1,\ldots,k}R(y_i,x)
\right]
+c(x).
\end{equation}
The shift therefore adds a policy-independent constant to
Eqs.~\ref{eq:single_reward_objective}
and~\ref{eq:k_sample_objective}, leaving their optimal policies
unchanged. Consequently, if rewards are bounded below by
$r_{\min}$, we may replace $R$ with $R-r_{\min}$ and work with
nonnegative rewards without loss of generality.

\paragraph{Nonnegative rewards and effective regularization.}
Suppose that the reward support for prompt $x$ has lower endpoint
$a_x$. Since $F_{\pi^*}(u\mid x)=0$ below $a_x$, the exponent in
Eq.~\ref{eq:maxk_optimal_policy_implicit} can be written, for
$r>a_x$, as
\begin{equation}
\frac{k}{\beta}
\int_{a_x}^{r}
F_{\pi^*}(u\mid x)^{k-1}\,du
=
\frac{r-a_x}{\beta_{\mathrm{eff}}(r,x)},
\end{equation}
where
\begin{equation}
\beta_{\mathrm{eff}}(r,x)
=
\frac{\beta(r-a_x)}
{k\displaystyle\int_{a_x}^{r}
F_{\pi^*}(u\mid x)^{k-1}\,du}.
\end{equation}
Shifting the lower endpoint to zero gives the simpler expression in
Eq.~\ref{eq:maxk_effective_beta}. The nonnegative convention therefore
provides a natural origin relative to which reward magnitude and
effective regularization can be compared. Without such an origin, a
definition based directly on $r/\beta_{\mathrm{eff}}$ would depend on
an arbitrary reward shift and could become singular or negative.

This convention is not required for the transformed-reward
interpretation. More generally, for any anchor $a$,
\begin{equation}
S_{x,a}(r)
=
k\int_a^r
F_{\pi^*}(u\mid x)^{k-1}\,du.
\end{equation}
Changing $a$ adds the same constant to $S_{x,a}(r)$ for every
completion of a fixed prompt. This constant is absorbed into the
partition function and therefore does not affect the optimal policy.

\paragraph{Common reward range.}
Our cross-prompt interpretation treats the location of the reference
reward distribution --- and, informally, its mean
\begin{equation}
\mu_{\mathrm{ref}}(x)
=
\mathbb{E}_{y\sim\pi_{\mathrm{ref}}(\cdot\mid x)}
[R(y,x)]
\end{equation}
--- as an indicator of task difficulty: prompts receiving lower rewards
under the reference policy are regarded as more difficult. This
interpretation is meaningful only when rewards can be compared across
prompts. In particular, a reward value $r_1$ must represent a better
outcome than $r_2<r_1$ independently of which prompts produced the two
values.

We therefore assume a shared, semantically meaningful fixed range,
\begin{equation}
R(y,x)\in[r_{\min},r_{\max}]
\qquad\text{for every }x.
\end{equation}
Without common anchors, the average reward cannot define prompt
difficulty consistently. Indeed, because the objectives are invariant
to prompt-dependent reward shifts, one could change the mean reward of
any prompt arbitrarily without changing either its optimization
problem or its optimal policy. A difference between two means would
then reflect only a difference in reward coordinates rather than a
difference in task difficulty. Prompt-dependent rescalings create the
same comparability problem and additionally alter the balance between
reward and KL regularization unless $\beta$ is rescaled accordingly.
A common fixed range rules out these ambiguities, as in normalized
verifier scores and test-case pass rates.

The mean provides an intuitive summary of difficulty but does not, by
itself, guarantee a pointwise ordering of the adaptive transformations.
A sufficient condition for the ordering illustrated in
Figures~\ref{fig:maxk_effective_beta}
and~\ref{fig:maxk_reward_transform} can be stated in terms of the
optimal reward CDFs. Consider an easy prompt $x_{\mathrm{e}}$ and a difficult
prompt $x_{\mathrm{h}}$ on a common range $[0,r_{\max}]$ such that
\begin{equation}
F_{\pi^*}(r\mid x_{\mathrm{e}})
\leq
F_{\pi^*}(r\mid x_{\mathrm{h}})
\qquad
\text{for all }r\in[0,r_{\max}].
\end{equation}
Then, for every common raw reward value,
\begin{equation}
S_{x_{\mathrm{e}}}(r)
\leq
S_{x_{\mathrm{h}}}(r),
\qquad
\beta_{\mathrm{eff}}(r,x_{\mathrm{e}})
\geq
\beta_{\mathrm{eff}}(r,x_{\mathrm{h}}).
\end{equation}
Thus, the easy prompt receives a smaller transformed reward and
stronger effective regularization. This condition formalizes the
representative easy--difficult comparison used in the main text; it is
not a universal definition of task difficulty. If the reward CDFs
cross, the corresponding transformation and effective-regularization
curves may also cross, and no global ordering is implied.

\section{Explicit \texorpdfstring{$\mathrm{max@k}$}{max@k} Solution for Win-Rate Rewards}
\label{app:winrate_maxk_derivation}

We now derive the explicit solution in Eqs.~\ref{eq:quantile_maxk_optimal_policy}--\ref{eq:quantile_maxk_transform}. Fix a prompt $x$ and suppress its dependence where unambiguous. Let
\begin{equation}
F_{\mathrm{ref}}(r\mid x)
=
\mathbb{P}_{y'\sim\pi_{\mathrm{ref}}(\cdot\mid x)}
[R(y',x)\leq r]
\end{equation}
denote the reference reward CDF, and define the reference quantile, or win rate, of a completion $y$ as
\begin{equation}
Q(y)=F_{\mathrm{ref}}(R(y,x)\mid x).
\end{equation}

\noindent\textbf{Proposition~\ref{prop:winrate_maxk_optimal_policy} (restated).}
Suppose that the reference reward distribution is continuous. For $k\geq1$ and $\beta>0$, the KL-regularized $\mathrm{max@k}$ objective with reward $Q(y)$ has the unique optimal policy
\[
\pi_q^*(y\mid x)
=
\pi_{\mathrm{ref}}(y\mid x)
\exp\left(\frac{S_x^{\mathrm q}(R(y,x))}{\beta}\right),
\]
where
\[
S_x^{\mathrm q}(r)
=
\beta\log\left(
C+\frac{[\Psi^{-1}(F_{\mathrm{ref}}(r\mid x))]^k}{\beta}
\right),
\qquad
\Psi(z)=\int_0^z\frac{du}{C+u^k/\beta},
\]
and $C>0$ is uniquely determined by $\Psi(1)=1$. The partition function for this transformed reward is exactly one.

\begin{proof}
\textit{Step 1: Deriving the optimal win-rate density.}
Since the reference reward distribution is atomless, the probability integral transform gives
\begin{equation}
Q(y)\sim\mathrm{Unif}[0,1],
\qquad
y\sim\pi_{\mathrm{ref}}(\cdot\mid x).
\end{equation}
Let
\begin{equation}
G(q)
=
\mathbb{P}_{y\sim\pi_q^*(\cdot\mid x)}
[Q(y)\leq q]
\end{equation}
be the win-rate CDF under the optimal policy, and let $g(q)=G'(q)$ denote its density. The optimality condition from Appendix~\ref{app:maxk_derivation}, applied with $Q(y)$ as the reward, gives
\[
\pi_q^*(y\mid x)=\pi_{\mathrm{ref}}(y\mid x)\,w(Q(y)),
\qquad
w(q)=\frac{1}{Z(x)}
\exp\left(\frac{k}{\beta}\int_0^q G(u)^{k-1}\,du\right).
\]
For $k=1$, the same expression follows directly from the ordinary KL-regularized optimal policy in Eq.~\ref{eq:optimal_policy}.

The multiplier $w(Q(y))$ depends on the completion only through its win rate. Since $Q$ is uniform under the reference policy, its CDF under the reweighted policy satisfies
\[
G(q)
=
\mathbb{E}_{y\sim\pi_{\mathrm{ref}}(\cdot\mid x)}
\left[w(Q(y))\,\mathbb{I}\{Q(y)\leq q\}\right]
=
\int_0^q w(u)\,du.
\]
Differentiating gives $g(q)=w(q)$, and hence
$\pi_q^*(y\mid x)=\pi_{\mathrm{ref}}(y\mid x)\,g(Q(y))$.
In particular,
\begin{equation}\label{eq:app_quantile_density_implicit}
g(q)
=
\frac{1}{Z(x)}
\exp\left(
\frac{k}{\beta}\int_0^q G(u)^{k-1}\,du
\right).
\end{equation}

Differentiating Eq.~\ref{eq:app_quantile_density_implicit} yields
\begin{equation}
\frac{g'(q)}{g(q)}
=
\frac{k}{\beta}G(q)^{k-1}.
\end{equation}
Using $G'(q)=g(q)>0$ and treating $g$ as a function of $G$, we obtain
\begin{equation}
\frac{dg}{dG}
=
\frac{g'(q)}{G'(q)}
=
\frac{k}{\beta}G^{k-1}.
\end{equation}
Integration therefore gives
\begin{equation}\label{eq:app_quantile_density_explicit}
g(q)=C+\frac{1}{\beta}G(q)^k,
\end{equation}
where $C=g(0)>0$. Thus, the optimal win-rate CDF satisfies
\begin{equation}
G'(q)=C+\frac{1}{\beta}G(q)^k,
\qquad
G(0)=0,
\qquad
G(1)=1.
\end{equation}

\textit{Step 2: Determining the constant and the CDF.}
Separating variables gives
\begin{equation}
q
=
\int_0^{G(q)}\frac{du}{C+u^k/\beta}
=
\Psi(G(q)).
\end{equation}
The terminal condition $G(1)=1$ requires
\begin{equation}\label{eq:app_quantile_constant}
1=\int_0^1\frac{du}{C+u^k/\beta}.
\end{equation}
The right-hand side is continuous and strictly decreasing in $C>0$. It diverges as $C\downarrow0$ for every $k\geq1$ and converges to zero as $C\to\infty$. Hence Eq.~\ref{eq:app_quantile_constant} has a unique solution $C>0$. For this value of $C$, $\Psi$ is a strictly increasing bijection from $[0,1]$ to $[0,1]$, and therefore
\begin{equation}
G=\Psi^{-1}.
\end{equation}

\textit{Step 3: Recovering and verifying the optimal policy.}
As established in Step 1, the optimal completion policy satisfies $\pi_q^*(y\mid x)=\pi_{\mathrm{ref}}(y\mid x)\,g(Q(y))$. Substituting the density from Eq.~\ref{eq:app_quantile_density_explicit} and $Q(y)=F_{\mathrm{ref}}(R(y,x)\mid x)$ gives
\begin{equation}
\pi_q^*(y\mid x)
=
\pi_{\mathrm{ref}}(y\mid x)
\left(
C+\frac{[G(F_{\mathrm{ref}}(R(y,x)\mid x))]^k}{\beta}
\right).
\end{equation}
Equivalently, defining
\begin{equation}
S_x^{\mathrm q}(r)
=
\beta\log\left(
C+\frac{[G(F_{\mathrm{ref}}(r\mid x))]^k}{\beta}
\right)
\end{equation}
yields
\begin{equation}
\pi_q^*(y\mid x)
=
\pi_{\mathrm{ref}}(y\mid x)
\exp\left(\frac{S_x^{\mathrm q}(R(y,x))}{\beta}\right).
\end{equation}
Normalization is already incorporated because
\begin{equation}
\mathbb{E}_{y\sim\pi_{\mathrm{ref}}(\cdot\mid x)}
\left[
\exp\left(\frac{S_x^{\mathrm q}(R(y,x))}{\beta}\right)
\right]
=
\int_0^1g(q)\,dq
=
G(1)-G(0)
=
1.
\end{equation}

To verify sufficiency, observe that the constructed policy induces win-rate density $g$ and hence CDF $G$. Moreover, differentiating $g=C+G^k/\beta$ with $G'=g$ gives $g'/g=(k/\beta)G^{k-1}$. Integrating and using $g(0)=C$ recovers Eq.~\ref{eq:app_quantile_density_implicit}, with normalizing constant $1/C$. The policy therefore satisfies the sufficient optimality condition from Appendix~\ref{app:maxk_derivation}. Strict concavity of the KL-regularized objective establishes uniqueness.
\end{proof}

The quantities $C$, $\Psi$, and $G$ depend only on $k$ and $\beta$ and can therefore be precomputed independently of the prompt. All prompt dependence enters through the reference CDF $F_{\mathrm{ref}}$.

For $k=1$, the construction reduces to the standard exponential tilt:
\begin{equation}
g(q)
=
\frac{\exp(q/\beta)}
{\displaystyle\int_0^1\exp(u/\beta)\,du}.
\end{equation}

\paragraph{Ties and discrete rewards.}
In practical settings such as test-case pass-rate rewards, the reference reward distribution is discrete and contains ties. The deterministic win rate $Q(y)=F_{\mathrm{ref}}(R(y,x)\mid x)$ is then not exactly uniform under the reference policy, so the continuous solution provides an approximation to the optimal policy for this win-rate objective. We retain this deterministic convention in our experiments and evaluate the resulting adaptation empirically. The identity $Z(x)=1$ likewise holds for the true CDF of a continuous reference reward distribution, but need not hold for discrete rewards or finite-bank CDF estimates. Our implementation retains $Z(x)=1$ as part of the same continuous-reference approximation.

\section{\textsc{KAPPO} Adaptation of GRPO}
\label{app:grpo_adaptation}

GRPO constructs its learning signal by centering and standardizing
rewards within each prompt's completion group. This section explains
how that normalization interacts with the adaptive reward
transformations derived in Section~\ref{sec:optimal_policies}.
For a group of rewards $R_1,\ldots,R_n$, write
$\bar R=\frac{1}{n}\sum_i R_i$ and
$s_R=\operatorname{std}_i(R_i)$.
For clarity, we omit the numerical denominator stabilizer and write
the expressions below for nonzero-variance groups; constant-reward
groups receive zero advantages.

\paragraph{Cancellation of binary reward scaling.}
For binary rewards, Corollary~\ref{cor:passk_adaptive_reward} gives
the transformed reward $\widetilde R_i=a_k(x)R_i$, where
\begin{equation}
a_k(x)
=
\frac{\beta}{\beta_{\mathrm{eff}}(x)}
=
k(1-p^*(x))^{k-1}.
\end{equation}
Since this multiplier is shared by all completions of the same
prompt, it scales both the centered rewards and their standard
deviation. For $a_k(x)>0$, the resulting advantages satisfy
\begin{equation}\label{eq:grpo_binary_cancellation}
\frac{\widetilde R_i-\overline{\widetilde R}}
{\operatorname{std}_j(\widetilde R_j)}
=
\frac{a_k(x)(R_i-\bar R)}{a_k(x)s_R}
=
\frac{R_i-\bar R}{s_R}.
\end{equation}
Thus, substituting the transformed binary rewards before group
standardization cancels the intended prompt-dependent scaling.

\paragraph{Preserving the binary adaptation.}
The core issue here is that standardized GRPO does not generally estimate the policy gradient of the underlying expected-reward objective. Dividing each group by its empirical reward standard deviation reweights each prompt's update by a factor that depends on the sampled completions and the current policy; consequently, the expected GRPO update generally differs from the policy gradient of the original objective. For the same reason, KAPPO's adaptive reward scaling is discarded, as shown in Eq.~\ref{eq:grpo_binary_cancellation}. Therefore, the failure of ordinary reward substitution in this case arises precisely because standardized GRPO does not follow the objective to which that connection applies.

Accordingly, we treat the KAPPO adaptation of standardized GRPO as an empirical extension of our theoretical result. To retain the derived redistribution of pressure while preserving GRPO's normalization, we apply the adaptive multiplier after standardization: \begin{equation}\label{eq:grpo_post_normalization_scaling}
\widehat A_i^k
=
a_k(x)\frac{R_i-\bar R}{s_R}.
\end{equation}
Since the clipped GRPO surrogate is positively homogeneous in the advantage, this construction multiplies each prompt's contribution to the GRPO loss and gradient by exactly $a_k(x)$. It therefore preserves the theoretically derived cross-prompt pressure redistribution within GRPO's existing learning signal, without claiming an exact objective-level equivalence. This provides a natural reward-side adaptation for GRPO without KL regularization.

Two alternative implementations recover the theoretically grounded interfaces more directly. Removing standard-deviation normalization while retaining group centering allows the ordinary KAPPO reward scaling to survive. Alternatively, for KL-regularized GRPO, the regularization-side interface can be applied through the adaptive coefficient $\beta_{\mathrm{eff}}(x)$, as established in Corollary~\ref{cor:passk_adaptive_regularization}.

\paragraph{Continuous reward transformations.}
For continuous rewards, group standardization does not cause the same cancellation. The transformation $S_x^{\mathrm q}$ is nonlinear for $k>1$, whereas centering and standardization eliminate only affine changes in reward. The transformed rewards are therefore affected by GRPO's normalization but are not generally discarded by it. This is consistent with the purpose of the adaptation: KAPPO supplies the $\mathrm{@k}$ reward transformation, while GRPO retains its defining group-relative reweighting of the resulting learning signal.

For the win-rate-based transformation used in our experiments, the interaction is particularly clear. At the beginning of training, when the current policy is close to the reference policy, the population mean and standard deviation of the transformed rewards are approximately the same for every prompt. GRPO normalization is therefore trivial with respect to prompt adaptivity: it applies the same affine transformation across prompts and does not alter the redistribution of optimization pressure induced by KAPPO.

Indeed, write
\begin{equation}
Q_x(Y)
=
F_{\mathrm{ref}}(R(Y,x)\mid x),
\qquad
S_x^{\mathrm q}(R(Y,x))
=
T_{k,\beta}(Q_x(Y)),
\end{equation}
where $T_{k,\beta}$ is shared across prompts. Under the continuous-reference assumption from Appendix~\ref{app:winrate_maxk_derivation}, at initialization
\begin{equation}
Q_x(Y)\sim\operatorname{Unif}[0,1],
\qquad
Y\sim\pi_{\mathrm{ref}}(\cdot\mid x).
\end{equation}
Hence,
\begin{equation}
\mu_{k,\beta}
=
\mathbb E_{U\sim\operatorname{Unif}[0,1]}[T_{k,\beta}(U)],
\qquad
\sigma_{k,\beta}^2
=
\operatorname{Var}_{U\sim\operatorname{Unif}[0,1]}
[T_{k,\beta}(U)]
\end{equation}
do not depend on the prompt, and GRPO applies the common normalization
\begin{equation}
S_x^{\mathrm q}(r)
\longmapsto
\frac{S_x^{\mathrm q}(r)-\mu_{k,\beta}}
{\sigma_{k,\beta}}.
\end{equation}
Thus, the prompt-adaptive structure of $S_x^{\mathrm q}$ is preserved up to a common shift and scale.

As training proceeds, the current policy's win-rate distribution can evolve differently across prompts, so the corresponding means and standard deviations become prompt dependent. GRPO normalization then becomes nontrivial and interacts with the KAPPO transformation, introducing GRPO's usual prompt-dependent reweighting on top of the adaptive pressure supplied by KAPPO. Since $S_x^{\mathrm q}$ is nonlinear, this normalization affects but does not cancel the $\mathrm{@k}$ transformation.

\section{Experimental Details}
\label{app:experimental_details}

\paragraph{General setup.}
We perform full-parameter training in bfloat16 using AdamW with zero
weight decay. We use a cosine learning-rate schedule with a $3\%$
warmup, a maximum prompt and response length of 4096 tokens, and sample
training completions with temperature $1$ and top-$p=1$. Hyperparameters
and intermediate checkpoints are selected using the corresponding
validation sets. All \textsc{KAPPO} adaptations use $M=30$ reference completions
per prompt, generated once before training.

\paragraph{Mathematical reasoning.}
We train for at most 1000 steps. For
$\mathrm{@k}$-$A^{\star}$PO, we search learning rates in
$\{5\mathord{\times}10^{-6},10^{-5},2\mathord{\times}10^{-5}\}$ and
$\beta$ in
$\{2.5\mathord{\times}10^{-4},5\mathord{\times}10^{-4},10^{-3}\}$.
For PKPO, we search learning rates in
$\{10^{-6},5\mathord{\times}10^{-6},10^{-5}\}$ and use the minimum
rollout multiplicity $n=k$. GRPO uses $n=8$ completions per prompt and
learning rates in
$\{5\mathord{\times}10^{-6},10^{-5},2\mathord{\times}10^{-5}\}$.
It uses standard group-relative advantage normalization without an
additional KL penalty.

For standard DPO, we search learning rates in
$\{5\mathord{\times}10^{-6},10^{-5},2\mathord{\times}10^{-5}\}$ and
$\beta$ in $\{0.01,0.02,0.05\}$. For
$\mathrm{@k}$-DPO, we use the same learning-rate grid and search the
base coefficient used to construct $\beta_{\mathrm{eff}}(x)$ in
$\{10^{-4},2\mathord{\times}10^{-4},5\mathord{\times}10^{-4}\}$.
Both variants construct a preference pair by sampling eight
completions and selecting the highest- and lowest-reward candidates;
ties receive zero loss weight. We use 64 prompts per DPO step, yielding
128 selected completions after pair construction, and disable
log-probability length normalization. For $\mathrm{@k}$-DPO, we add a failure pseudo-count of $\alpha=0.5$ and cap $\beta_{\mathrm{eff}}(x)$ at $2$; the precise boundary handling is described in Appendix~\ref{app:calibration_boundaries}. The other mathematical-reasoning
experiments use 128 prompts per training step.

\paragraph{Code generation.}
We train Qwen3-4B for at most 500 steps using 128 prompts per step.
For win-rate $\mathrm{@k}$-$A^{\star}$PO, we search learning rates in
$\{10^{-5},2\mathord{\times}10^{-5}\}$ and $\beta$ in
$\{5\mathord{\times}10^{-4},10^{-3},5\mathord{\times}10^{-3}\}$.
For PKPO and GRPO, we search learning rates in
$\{10^{-6},5\mathord{\times}10^{-6},10^{-5}\}$. PKPO uses $n=k$,
whereas GRPO uses $n=8$. During training, rewards are computed using up
to 64 held-out test cases per problem; validation and final evaluation
use the complete available test suite.

For win-rate $\mathrm{@k}$-GRPO, we search learning rates in $\{5\mathord{\times}10^{-6},10^{-5}\}$ and the reward-transformation parameter $\beta$ in $\{5\mathord{\times}10^{-4},10^{-3},5\mathord{\times}10^{-3}\}$. As in standard GRPO, we use $n=8$ completions per prompt and no KL penalty.

\section{Compute Resources}
\label{app:compute_resources}

All training runs used 32 NVIDIA GH200 GPUs. Both mathematical
reasoning and code-generation runs took approximately 10--12 hours, corresponding to 320--384 GPU-hours per run. The overall computational cost of the study, including hyperparameter sweeps, exploratory runs, and evaluation exceeded 100{,}000 GPU-hours.

\section{Reference Calibration}
\label{app:reference_calibration}

\textsc{KAPPO} requires prompt-level statistics of the reference
policy. We estimate them once before training from a fixed bank of
$M$ reference completions per prompt. Since the reference policy is
fixed, the entire bank can be generated as an inference-only job using
high-throughput engines such as vLLM or SGLang. Generation can be
continuously batched and performed asynchronously, without backward
passes, optimizer updates, parameter offloading, or repeated weight
synchronization between the actor and generation workers. The
resulting rewards are stored and reused across all training runs that
share the same starting model and dataset.

For binary rewards, calibration estimates
\begin{equation}
\widehat p_{\mathrm{ref}}(x)
=
\frac{1}{M}\sum_{j=1}^{M}
\mathbb I[R(y_j^{\mathrm{ref}},x)=1].
\end{equation}
For each choice of $k$ and $\beta$, we then solve
Eq.~\ref{eq:passk_fixed_point} for $p^*(x)$ and compute
$\beta_{\mathrm{eff}}(x)$ and $Z(x)$ (defined in Eq.~\ref{eq:passk_effective_beta}). The
$\mathrm{@k}$-DPO instantiation uses $\beta_{\mathrm{eff}}(x)$,
whereas binary $\mathrm{@k}$-$A^{\star}$PO additionally uses $Z(x)$
and the reward multiplier
$\beta/\beta_{\mathrm{eff}}(x)$.

For continuous rewards, the same bank defines the empirical reference
CDF
\begin{equation}
\widehat F_{\mathrm{ref}}(r\mid x)
=
\frac{1}{M}\sum_{j=1}^{M}
\mathbb I[R(y_j^{\mathrm{ref}},x)\leq r].
\end{equation}
It is substituted into the explicit transformation
$S_x^{\mathrm q}$ in
Eq.~\ref{eq:quantile_maxk_transform}. The remaining quantities
$C$ and $\Psi$ depend only on $k$ and $\beta$ and are precomputed
once. Both continuous $\mathrm{@k}$-$A^{\star}$PO and
$\mathrm{@k}$-GRPO therefore reuse exactly the same calibrated
reference rewards; changing $k$, $\beta$, the learning rate, or the
random seed requires no additional reference generation.

We use $M=30$ throughout our experiments. For MATH, this corresponds
to $30\times7{,}500=225{,}000$ reference completions per model.
Generating the complete bank took approximately $0.5$ hours in our
setup, whereas a single 1,000-step training run took approximately
$10$--$12$ hours. Thus, even if charged entirely to one run,
calibration contributes only
\begin{equation}
\frac{0.5}{10\text{--}12}
\approx 4.2\%\text{--}5.0\%
\end{equation}
of its wall-clock time. More importantly, this is a one-time cost
shared by every run using the same model and dataset. Amortized over
$N$ runs, its relative contribution is only
$(4.2\%\text{--}5.0\%)/N$; over six runs, for example, it adds about
five minutes per run, or less than $1\%$. The bank is typically reused
across substantially more configurations, making the practical cost
of reference calibration negligible relative to training.

\subsection{Calibration Algorithms}
\label{app:calibration_algorithms}

Algorithms~\ref{alg:binary_calibration} and~\ref{alg:winrate_calibration} summarize reference calibration for binary and continuous win-rate rewards, respectively. Both start from the same reusable reference bank: for each training prompt $x\in\mathcal X$, draw $y_{x,1}^{\mathrm{ref}},\ldots,y_{x,M}^{\mathrm{ref}}\iid\pi_{\mathrm{ref}}(\cdot\mid x)$ and store their rewards $r_{x,j}^{\mathrm{ref}}=R(y_{x,j}^{\mathrm{ref}},x)$. This sampling is performed only once per reference model and dataset. Changing $k$ or $\beta$ requires only recomputing the numerical tables below, not generating additional completions.

\paragraph{Binary rewards.}
The calibration depends on each prompt only through its reference success count $N(x)$. We therefore precompute the required quantities for all $M+1$ possible counts. Algorithm~\ref{alg:binary_calibration} includes the method-specific boundary conventions used in our experiments: no smoothing for $A^{\star}$PO, and a failure pseudo-count with a coefficient cap for DPO. We store $\ell(x)=\beta\log\widehat Z(x)$ rather than $\widehat Z(x)$.

\begin{algorithm}[t]
\caption{Reference calibration for binary rewards}
\label{alg:binary_calibration}
\begin{algorithmic}[1]
\Require Reference rewards $\{r_{x,j}^{\mathrm{ref}}\}$, bank size $M$, training parameter $k\geq1$, $\beta>0$, method $m\in\{A^{\star}\mathrm{PO},\mathrm{DPO}\}$
\State $\alpha\gets0$ for $A^{\star}$PO, or $\alpha\gets0.5$ for DPO
\For{$j=0,\ldots,M$}
    \State $q_j\gets j/(M+\alpha)$
    \If{$q_j=0$}
        \State $p_j\gets0$; $a_j\gets k$; $\ell_j\gets0$
    \ElsIf{$q_j=1$}
        \State $p_j\gets1$
        \State $a_j\gets1$ if $k=1$, otherwise $a_j\gets0$
        \State $\ell_j\gets a_j$
    \Else
        \State Solve Eq.~\ref{eq:passk_fixed_point} with $p_{\mathrm{ref}}=q_j$ for $p_j$
        \State $a_j\gets k(1-p_j)^{k-1}$
        \State $\ell_j\gets\beta\log\!\left(1-q_j+q_j\exp(a_j/\beta)\right)$
    \EndIf
    \If{$m=\mathrm{DPO}$}
        \State $b_j\gets\min\{\beta/a_j,\,2\}$
    \EndIf
\EndFor
\For{each prompt $x\in\mathcal X$}
    \State $N(x)\gets\sum_{j=1}^{M}\mathbb I[r_{x,j}^{\mathrm{ref}}=1]$
    \State $\widehat p^*(x)\gets p_{N(x)}$
    \If{$m=A^{\star}\mathrm{PO}$}
        \State $\widehat a_k(x)\gets a_{N(x)}$; $\ell(x)\gets\ell_{N(x)}$
    \Else
        \State $\widehat\beta_{\mathrm{eff}}(x)\gets b_{N(x)}$
    \EndIf
\EndFor
\State \Return $\{\widehat a_k(x),\ell(x)\}_{x\in\mathcal X}$ for $A^{\star}$PO, or $\{\widehat\beta_{\mathrm{eff}}(x)\}_{x\in\mathcal X}$ for DPO
\end{algorithmic}
\end{algorithm}

During training, binary $\mathrm{@k}$-$A^{\star}$PO uses the regression target $\widehat a_k(x)R(y,x)-\ell(x)$, while $\mathrm{@k}$-DPO uses $\widehat\beta_{\mathrm{eff}}(x)$ in its preference loss. Neither requires further calibration during training. The fixed-point solve and partition computation are implemented in numerically stable coordinates, as described in Appendix~\ref{app:calibration_boundaries}.

\paragraph{Continuous win-rate rewards.}
An empirical CDF constructed from $M$ reference rewards takes values only in $\{0,1/M,\ldots,1\}$. Consequently, the entire win-rate-based transformation can be precomputed on this grid. The resulting table is shared by all prompts and by both continuous $\mathrm{@k}$-$A^{\star}$PO and $\mathrm{@k}$-GRPO.

\begin{algorithm}[t]
\caption{Reference calibration for continuous win-rate rewards}
\label{alg:winrate_calibration}
\begin{algorithmic}[1]
\Require Reference rewards $\{r_{x,j}^{\mathrm{ref}}\}$, bank size $M$, training parameter $k\geq1$, $\beta>0$
\If{$k=1$}
    \State $\log C\gets-\log\beta-\beta^{-1}-\log(1-e^{-1/\beta})$
    \For{$j=0,\ldots,M$}
        \State $T_j\gets j/M+\beta\log C$
    \EndFor
\Else
    \State Solve $\displaystyle\int_0^1\frac{du}{C+u^k/\beta}=1$ for $C>0$
    \State Define $\displaystyle\Psi(z)=\int_0^z\frac{du}{C+u^k/\beta}$
    \State $T_0\gets\beta\log C$; $T_M\gets\beta\log(C+1/\beta)$
    \For{$j=1,\ldots,M-1$}
        \State Solve $\Psi(z_j)=j/M$ for $z_j\in(0,1)$
        \State $T_j\gets\beta\log(C+z_j^k/\beta)$
    \EndFor
\EndIf
\State \Return Stored reference rewards and the shared table $\{T_j\}_{j=0}^{M}$
\Statex
\Function{TransformReward}{$x,r$}
    \State $j\gets\sum_{i=1}^{M}\mathbb I[r_{x,i}^{\mathrm{ref}}\leq r]$
    \State \Return $T_j$
\EndFunction
\end{algorithmic}
\end{algorithm}

For an online completion with reward $r$, \textsc{TransformReward} returns $\widehat S_x^{\mathrm q}(r)=T_{M\widehat F_{\mathrm{ref}}(r\mid x)}$ using only comparisons with stored rewards and a table lookup. No numerical integration or root finding is performed during training. The inclusive comparison preserves our convention for ties and includes both empirical-CDF endpoints without smoothing.

Continuous $\mathrm{@k}$-$A^{\star}$PO uses this value as its regression target, retaining $Z(x)=1$ under the continuous-reference approximation discussed in Appendix~\ref{app:winrate_maxk_derivation}. $\mathrm{@k}$-GRPO uses the same transformed rewards before its usual group normalization. The parameter $\beta$ determines the transformation in both cases; using it for calibration does not require adding a KL penalty to GRPO.

\subsection{Boundary Handling}
\label{app:calibration_boundaries}

A finite reference bank can contain no successes or only successes even when the true success probability lies strictly between zero and one. The implementation therefore needs explicit conventions for these empirical endpoints, distinct from the population boundary cases in Appendix~\ref{app:passk_derivation}. Let $N(x)$ denote the number of successful completions among the $M$ reference samples.

\paragraph{Binary $\mathrm{@k}$-$A^{\star}$PO.}
We use the unsmoothed estimate $\widehat p_{\mathrm{ref}}(x)=N(x)/M$. Writing $a_k(x)=\beta/\beta_{\mathrm{eff}}(x)$, the endpoint values for $k>1$ are
\begin{equation}
\bigl(\widehat p^*(x),\widehat a_k(x),\beta\log\widehat Z(x)\bigr)
=
\begin{cases}
(0,k,0), & N(x)=0,\\
(1,0,0), & N(x)=M.
\end{cases}
\end{equation}
We compute the regression target $\widehat a_k(x)R(y,x)-\beta\log\widehat Z(x)$ directly, avoiding an explicit infinite $\beta_{\mathrm{eff}}(x)$ at $N(x)=M$. Thus, a zero-success bank gives target $kR(y,x)$, whereas an all-success bank gives target zero for $k>1$. For $k=1$, the multiplier remains one; the targets at $N(x)=0$ and $N(x)=M$ are $R(y,x)$ and $R(y,x)-1$, respectively. No probability smoothing is used for this method.

\paragraph{Binary $\mathrm{@k}$-DPO.}
Unlike the regression target, the DPO loss requires a finite coefficient. We therefore add a failure pseudo-count and solve the fixed-point equation using $\widehat p_{\mathrm{ref}}(x)=N(x)/(M+\alpha)$ with $\alpha=0.5$. This leaves the zero-success endpoint unchanged but moves the all-success estimate below one. The formula applies to every count, not only to all-success banks. We then use
\begin{equation}
\widehat\beta_{\mathrm{eff}}(x)
=\min\left\{\frac{\beta}{k(1-\widehat p^*(x))^{k-1}},\,2\right\}.
\end{equation}
Smoothing prevents the infinite coefficient at an empirical success probability of one, while the cap limits very large coefficients near this boundary.

\paragraph{Win-rate endpoints and ties.}
For continuous $\mathrm{@k}$-$A^{\star}$PO and $\mathrm{@k}$-GRPO, we use the inclusive empirical CDF without smoothing or clipping its values away from zero and one. The transformation remains finite at both endpoints:
\begin{equation}
S_x^{\mathrm q}(r)=
\begin{cases}
\beta\log C, & \widehat F_{\mathrm{ref}}(r\mid x)=0,\\
\beta\log(C+1/\beta), & \widehat F_{\mathrm{ref}}(r\mid x)=1.
\end{cases}
\end{equation}
Rewards below the smallest reference reward receive the first value; rewards at or above the largest receive the second. Ties retain the inclusive-CDF convention, including for a constant-reward reference bank. For $A^{\star}$PO, we retain $Z(x)=1$ under the continuous-reference approximation discussed in Appendix~\ref{app:winrate_maxk_derivation}.

\paragraph{Numerical stability.}
For interior binary estimates, we solve the scalar equation in log-odds coordinates and compute $a_k$ directly from that solution, avoiding loss of precision in $1-p^*$ near one. Partition terms and win-rate transformations are evaluated in the log domain, with a separate closed-form computation for $k=1$, to avoid overflow at small $\beta$. These numerical evaluations do not introduce additional probability smoothing.

\paragraph{Degenerate online groups.}
In $\mathrm{@k}$-GRPO, we divide centered transformed rewards by their group standard deviation plus $10^{-6}$; constant-reward groups therefore have zero advantages. For both DPO variants, selected pairs with equal rewards receive zero loss weight.

\section{Sensitivity to Finite Reference Calibration}
\label{app:calibration_sensitivity}

Reference calibration computes the quantities and mappings required by \textsc{KAPPO} before training. For binary rewards, we first estimate $p_{\mathrm{ref}}(x)$ from a fixed bank of reference completions, then solve Eq.~\ref{eq:passk_fixed_point} for $p^*(x)$ and compute the adaptive reward multiplier $a_k(x)$ or KL coefficient $\beta_{\mathrm{eff}}(x)$. For continuous rewards, we estimate the reference CDF $F_{\mathrm{ref}}(r\mid x)$ and substitute it into the win-rate-based transformation $S_x^{\mathrm q}(r)$. This section analyzes how finite-sample noise in the estimates of $p_{\mathrm{ref}}$ and $F_{\mathrm{ref}}$ propagates to these final quantities. We distinguish \emph{noise}, measured by the standard deviation across independently drawn reference banks, from \emph{bias}, measured by the difference between the mean estimate and the exact value obtained from the true reference distribution. The figures use our experimental bank size $M=30$ and an illustrative coefficient $\beta=0.1$.

\paragraph{Exact finite-sample calculation.}
Fix a prompt and let $q$ denote the true reference quantity being estimated: $q=p_{\mathrm{ref}}(x)$ for binary rewards, or $q=F_{\mathrm{ref}}(r\mid x)$ at a fixed reward threshold $r$ for continuous rewards. Each reference completion then produces a binary observation---success in the first case and the event $R(y,x)\leq r$ in the second. Hence, the number $N$ of positive observations in a bank of $M$ i.i.d. reference completions satisfies $N\sim\operatorname{Binomial}(M,q)$, and the empirical estimate is $\widehat q=N/M$.

Let $\theta(q)$ denote any exact downstream quantity computed from $q$, such as $p^*$, $a_k$, $\beta_{\mathrm{eff}}$, or $S_x^{\mathrm q}(r)$. Let $\widehat\theta_j$ be the value produced by the implemented calibration rule when $N=j$, including any smoothing or clipping. Since $\Pr(N=j)=w_j(q):=\binom{M}{j}q^j(1-q)^{M-j}$, its exact finite-bank mean, noise, and bias are
\begin{equation}\label{eq:calibration_exact_moments}
\mu_\theta(q)=\sum_{j=0}^{M}w_j(q)\widehat\theta_j,
\qquad
\sigma_\theta(q)=\left(\sum_{j=0}^{M}w_j(q)[\widehat\theta_j-\mu_\theta(q)]^2\right)^{1/2},
\qquad
b_\theta(q)=\mu_\theta(q)-\theta(q).
\end{equation}
We report absolute noise and bias for $p^*$ and $a_k$. Because the scale of $\beta_{\mathrm{eff}}$ varies substantially, for this quantity we define relative noise and relative bias as $\sigma_\theta(q)/\theta(q)$ and $b_\theta(q)/\theta(q)$, respectively, using the exact coefficient in the denominator. Equation~\ref{eq:calibration_exact_moments} sums over all $M+1$ possible reference-bank outcomes, without Monte Carlo simulation or a local approximation. The reported errors account for finite-bank estimation and, where applicable, smoothing and clipping; numerical solver error is not included in this analysis.

\subsection{Binary Rewards}

Suppressing the prompt argument, set $q=p_{\mathrm{ref}}(x)$ and let $g(q)$ be the optimal success probability determined by Eq.~\ref{eq:passk_fixed_point}. The exact maps applied during reference calibration are
\begin{equation}\label{eq:binary_calibration_maps}
p^*=g(q),
\qquad
a_k(q)=k[1-g(q)]^{k-1},
\qquad
\beta_{\mathrm{eff}}(q)=\frac{\beta}{a_k(q)},
\end{equation}
where $a_k$ is the adaptive reward multiplier. For $k>1$, the boundary values are $g(0)=0$, $g(1)=1$, $a_k(0)=k$, and $a_k(1)=0$. For $k=1$, the multiplier is identically one.

\begin{figure}[t]
    \centering
    \includegraphics[width=0.492\linewidth]{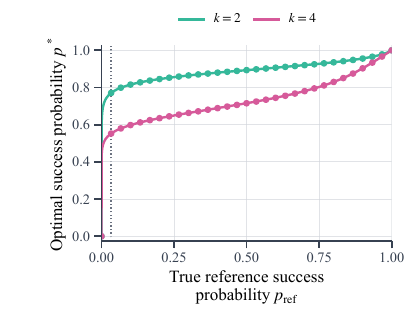}
    \caption{Exact optimal success probability as a function of the true reference success probability, for $\beta=0.1$. Markers show the attainable empirical estimates $j/M$ for $M=30$; the dotted line marks the first positive estimate $1/M$. The map is continuous, but its steep rise near zero is compressed on the linear horizontal scale.}
    \label{fig:calibration_binary_policy_map}
\end{figure}

\paragraph{Sensitive regions.}
Figure~\ref{fig:calibration_binary_policy_map} shows that $g$ rises sharply near zero and changes more gradually in the interior. A finite bank may therefore fail to resolve a narrow region over which the inferred $p^*$ changes substantially. For example, at $k=4$, $g(0)=0$ whereas $g(1/30)\approx0.553$, so observing one success instead of none produces a large change in $\widehat p^*$. The probability of observing no successes is $(1-q)^M$, which for $M=30$ is approximately $0.740$ when $q=0.01$ and $0.042$ when $q=0.1$.

\begin{figure}[t]
    \centering
    \begin{subfigure}{\linewidth}
        \centering
        \includegraphics[width=0.492\linewidth]{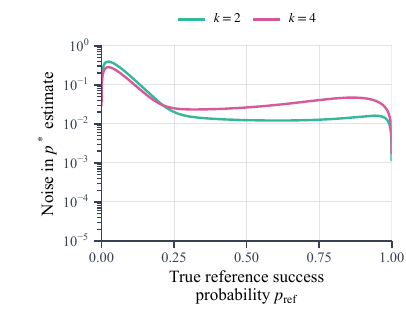}\hfill
        \includegraphics[width=0.492\linewidth]{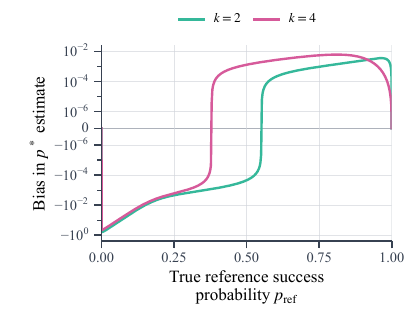}
        \caption{Optimal success probability $p^*$.}
    \end{subfigure}
    \par\smallskip
    \begin{subfigure}{\linewidth}
        \centering
        \includegraphics[width=0.492\linewidth]{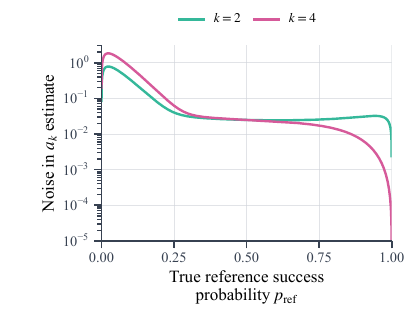}\hfill
        \includegraphics[width=0.492\linewidth]{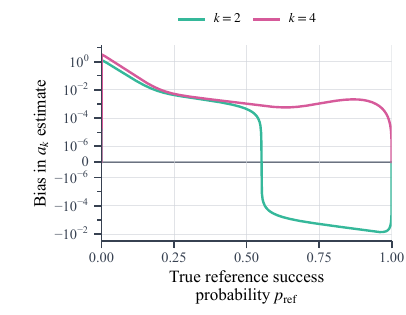}
        \caption{Adaptive reward multiplier $a_k$.}
    \end{subfigure}
    \par\smallskip
    \begin{subfigure}{\linewidth}
        \centering
        \includegraphics[width=0.492\linewidth]{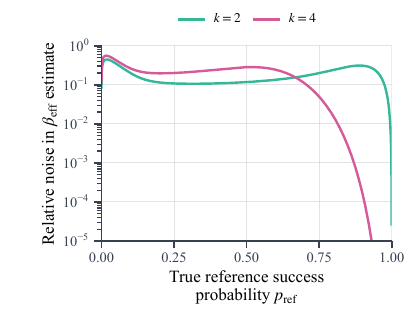}\hfill
        \includegraphics[width=0.492\linewidth]{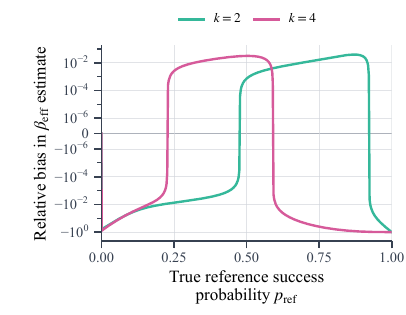}
        \caption{Adaptive KL coefficient $\beta_{\mathrm{eff}}$ (relative errors).}
    \end{subfigure}
    \caption{Exact finite-bank noise (left) and signed bias (right) for $M=30$ and $\beta=0.1$, plotted against the true reference success probability. Panels (a) and (b) use $\widehat q=N/M$; panel (c) uses Eq.~\ref{eq:calibration_dpo_estimator}. Coefficient errors are divided by the exact $\beta_{\mathrm{eff}}(q)$; the endpoint $q=1$, where this coefficient diverges for $k>1$, is omitted. Noise uses logarithmic axes, while signed bias uses symmetric-logarithmic axes that are linear within $[-10^{-6},10^{-6}]$. Zero noise at deterministic endpoints is omitted from the logarithmic axes.}
    \label{fig:calibration_binary_errors}
\end{figure}

\paragraph{Estimators used in the analysis.}
For $p^*$ and $a_k$, we use the unsmoothed estimate $\widehat q=N/M$ employed in our binary $\mathrm{@k}$-$A^{\star}$PO runs, giving $\widehat p^*=g(\widehat q)$ and $\widehat a_k=a_k(\widehat q)$. Thus, each horizontal-axis value in Figure~\ref{fig:calibration_binary_errors} fixes the true $p_{\mathrm{ref}}$, while the plotted noise describes how the inferred quantities vary across independently drawn reference banks. For $\beta_{\mathrm{eff}}$, we use the one-sided smoothing and cap employed by $\mathrm{@k}$-DPO:
\begin{equation}\label{eq:calibration_dpo_estimator}
\widehat q_{\mathrm D}=\frac{N}{M+0.5},
\qquad
\widehat\beta_{\mathrm{eff}}
=\min\left\{\frac{\beta}{a_k(\widehat q_{\mathrm D})},\,2\right\}.
\end{equation}
The smoothing keeps $\widehat q_{\mathrm D}<1$ while preserving $\widehat q_{\mathrm D}=0$ for a zero-success bank. Without smoothing or clipping, $N=M$ would produce an infinite coefficient for $k>1$, making its finite-bank moments undefined. We measure bias relative to the exact, unclipped $\beta_{\mathrm{eff}}(q)$, so the reported bias includes the effect of these implementation safeguards.

\paragraph{Consistency and convergence rates.}

\begin{lemma}[Consistency of binary calibration]
\label{lem:calibration_binary_consistency}
Fix an integer $k\geq1$, $\beta>0$, and $q\in[0,1]$. Under i.i.d. reference sampling, the unsmoothed estimates $\widehat p^*=g(\widehat q)$ and $\widehat a_k=a_k(\widehat q)$ are strongly consistent. For fixed $q\in(0,1)$ and $\theta\in\{g,a_k\}$, their noise and bias satisfy
\begin{equation}\label{eq:calibration_asymptotic_rates}
\sigma_\theta(q)
=
|\theta'(q)|\sqrt{\frac{q(1-q)}{M}}
+o(M^{-1/2}),
\qquad
b_\theta(q)
=
\frac{q(1-q)}{2M}\theta''(q)
+o(M^{-1}).
\end{equation}
At $q=0$ and $q=1$, both estimates are exact for every $M$.
\end{lemma}

\begin{proof}
By the strong law of large numbers, $\widehat q\to q$ almost surely. The fixed-point characterization in Eq.~\ref{eq:passk_fixed_point} implies that $g$ is continuous on $[0,1]$, including the limits $g(q)\to0$ as $q\downarrow0$ and $g(q)\to1$ as $q\uparrow1$. Thus, $a_k$ is also continuous, and applying either map to $\widehat q$ gives almost-sure convergence to its exact value.

For interior $q$, smoothness of $g$ follows from the implicit function theorem. Indeed, the derivative of the right-hand side minus the left-hand side of Eq.~\ref{eq:passk_fixed_point}, with respect to the candidate success probability $p$, is
\[
\frac{\beta}{p(1-p)}
+
k(k-1)(1-p)^{k-2}
>0.
\]
Consequently, both $g$ and $a_k$ are smooth on $(0,1)$.

For either map $\theta$, a Taylor expansion around the fixed interior point $q$, together with
\[
\mathbb E[\widehat q-q]=0,
\qquad
\mathbb E[(\widehat q-q)^2]=\frac{q(1-q)}{M},
\]
gives
\[
\mathbb E[\theta(\widehat q)]-\theta(q)
=
\frac{\theta''(q)q(1-q)}{2M}+o(M^{-1}),
\qquad
\operatorname{Var}(\theta(\widehat q))
=
\frac{\theta'(q)^2q(1-q)}{M}+o(M^{-1}).
\]
These expansions remain valid despite possible boundary behavior of the derivatives: outside a fixed interior neighborhood of $q$, binomial deviations have exponentially small probability, while $g$ and $a_k$ are bounded. Taking the square root of the variance gives Eq.~\ref{eq:calibration_asymptotic_rates}. Finally, when $q\in\{0,1\}$, $\widehat q=q$ almost surely for every $M$, so both errors vanish.
\end{proof}

Thus, sampling noise is $O(M^{-1/2})$ and bias is $O(M^{-1})$ at each fixed interior reference success probability. The asymptotic constants depend on the local slope and curvature of the calibration maps, which can be large near sensitive regions. Consequently, these rates describe the large-bank limit but need not accurately predict the errors at $M=30$; the figures instead use the exact finite-bank calculation in Eq.~\ref{eq:calibration_exact_moments}.

\paragraph{Effect of coefficient stabilization.}

\begin{lemma}[Consistency under smoothing and clipping]
\label{lem:calibration_stabilized_beta}
Fix an integer $k\geq1$, $\beta>0$, and $q\in(0,1)$. The estimator in Eq.~\ref{eq:calibration_dpo_estimator} satisfies
\[
\widehat\beta_{\mathrm{eff}}
\xrightarrow[M\to\infty]{\mathrm{a.s.}}
\min\{\beta_{\mathrm{eff}}(q),2\}.
\]
If $\beta_{\mathrm{eff}}(q)<2$, it is consistent for the exact coefficient, with $O(M^{-1/2})$ noise and $O(M^{-1})$ bias. If $\beta_{\mathrm{eff}}(q)>2$, its noise vanishes and its bias relative to the exact coefficient converges to $2-\beta_{\mathrm{eff}}(q)$; the probability of obtaining an estimate below the cap decays exponentially in $M$. At an interior clipping threshold $\beta_{\mathrm{eff}}(q)=2$ with $k>1$, the estimator remains consistent, but its downward bias is of order $M^{-1/2}$.
\end{lemma}

\begin{proof}
The smoothed estimate can be written as
\[
\widehat q_{\mathrm D}
=
\frac{M}{M+0.5}\widehat q.
\]
It therefore converges almost surely to $q$, and its bias is
\[
\mathbb E[\widehat q_{\mathrm D}]-q
=
-\frac{q}{2M+1}
=
O(M^{-1}).
\]
Continuity of the capped coefficient at every interior $q$ gives the stated almost-sure limit. Since the estimator is bounded by two, its mean and variance also converge to those of this deterministic limit.

When $\beta_{\mathrm{eff}}(q)<2$, the cap is inactive throughout a neighborhood of $q$. In this neighborhood, the coefficient is a smooth function of $\widehat q_{\mathrm D}$, so the same Taylor argument as in Lemma~\ref{lem:calibration_binary_consistency} gives $O(M^{-1/2})$ noise and $O(M^{-1})$ bias. The smoothing changes the leading bias coefficient but not its order. Excursions outside this neighborhood have exponentially small probability and do not change these rates.

When $\beta_{\mathrm{eff}}(q)>2$, the estimator equals two throughout a neighborhood of $q$. Obtaining a value below two therefore requires a deviation of $\widehat q_{\mathrm D}$ from $q$ by a fixed positive amount, whose probability decays exponentially by binomial concentration.

Finally, suppose that $k>1$ and $\beta_{\mathrm{eff}}(q)=2$. The coefficient is smooth and strictly increasing at this interior point. The central limit theorem and the local form of the clipping operation give
\[
\sqrt M\bigl(\widehat\beta_{\mathrm{eff}}-2\bigr)
\xrightarrow{\mathrm d}
\beta_{\mathrm{eff}}'(q)\min\{Z,0\},
\qquad
Z\sim\mathcal N(0,q(1-q)).
\]
Binomial moment bounds and local smoothness also justify convergence of the first moments. Hence,
\[
\mathbb E[\widehat\beta_{\mathrm{eff}}]-2
=
-\beta_{\mathrm{eff}}'(q)
\sqrt{\frac{q(1-q)}{2\pi M}}
+o(M^{-1/2}),
\]
establishing the downward bias of order $M^{-1/2}$.
\end{proof}

The failure pseudo-count thus vanishes asymptotically, whereas the fixed coefficient cap does not. Dividing the noise and bias by the fixed, finite $\beta_{\mathrm{eff}}(q)$ preserves the corresponding rates for the relative errors.

\paragraph{Implications.}
For $p^*$ and $a_k$, finite-bank error is concentrated near the low-success boundary and becomes much smaller in the interior. At $k=4$ and $p_{\mathrm{ref}}=0.1$, $\widehat p^*$ has standard deviation $0.123$ and bias $-0.028$; at $p_{\mathrm{ref}}=0.5$, these decrease to $0.026$ and $0.001$. The same boundary events affect the adaptive multiplier in the opposite direction: a bank containing no successes yields $\widehat p^*=0$ and hence the maximal multiplier $\widehat a_k=k$, producing an upward bias in the pressure assigned to prompts with small but nonzero success probability. As $M\to\infty$, both noise and bias vanish, and the probability $(1-p_{\mathrm{ref}})^M$ of unresolved zero-success banks decays exponentially. This improvement need not be monotone in $M$: near the low-success boundary, larger banks can initially increase variability as successful completions begin to appear. Additional reference completions are therefore most valuable for hard prompts for which $Mp_{\mathrm{ref}}$ is small. At the exact boundary $p_{\mathrm{ref}}=0$, there is no estimation error: every bank contains zero successes and the maximal multiplier is the correct value.

For $k>1$, the effective KL coefficient has a different limitation near $p_{\mathrm{ref}}=1$. Because the exact $\beta_{\mathrm{eff}}$ diverges while the implemented estimator is capped, its sampling noise can be small simply because nearly all estimates have saturated at the cap, even though its downward bias relative to the exact coefficient is large. Increasing $M$ reduces finite-bank noise asymptotically but cannot remove this cap-induced bias; doing so requires changing the cap or the stabilization rule. Thus, larger reference banks address the main uncertainty on low-success prompts, whereas accuracy on extremely easy prompts is ultimately limited by the chosen coefficient stabilization.

\subsection{Continuous Rewards}

For continuous rewards, reference calibration estimates the CDF value at each reward threshold and substitutes it into the win-rate-based transformation. At a fixed prompt and raw reward $r$, write Eq.~\ref{eq:quantile_maxk_transform} as
\begin{equation}\label{eq:calibration_winrate_map}
S_x^{\mathrm q}(r)=T(q),
\qquad
q=F_{\mathrm{ref}}(r\mid x),
\qquad
T(q)=\beta\log\left(C+\frac{G(q)^k}{\beta}\right),
\quad G=\Psi^{-1}.
\end{equation}
Here, $N$ counts the reference completions whose rewards do not exceed $r$, so $N\sim\operatorname{Binomial}(M,q)$ and $\widehat q=N/M$. The estimated transformed reward is therefore $T(\widehat q)$, and its exact noise and bias follow from Eq.~\ref{eq:calibration_exact_moments} by setting $\theta(q)=T(q)$ and $\widehat\theta_j=T(j/M)$. This pointwise calculation requires no model of the full raw-reward distribution. To compare different $k$, we divide the plotted noise and bias by the transformation range $\Delta_T=T(1)-T(0)$.

\begin{figure}[t]
    \centering
    \begin{subfigure}{0.492\linewidth}
        \centering
        \includegraphics[width=\linewidth]{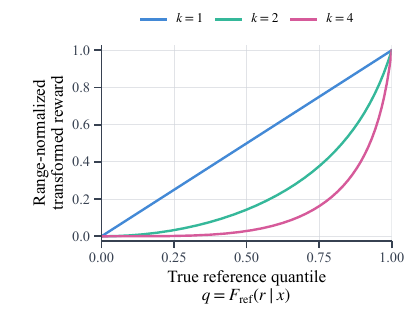}
        \caption{Win-rate-based reward transformation.}
    \end{subfigure}
    \par\smallskip
    \begin{subfigure}{0.492\linewidth}
        \centering
        \includegraphics[width=\linewidth]{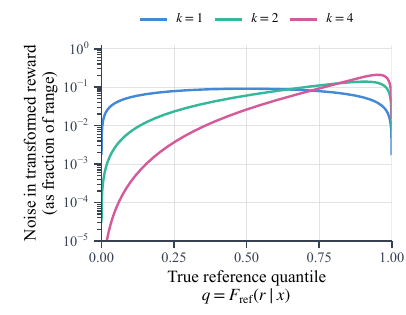}
        \caption{Noise in the transformed reward.}
    \end{subfigure}\hfill
    \begin{subfigure}{0.492\linewidth}
        \centering
        \includegraphics[width=\linewidth]{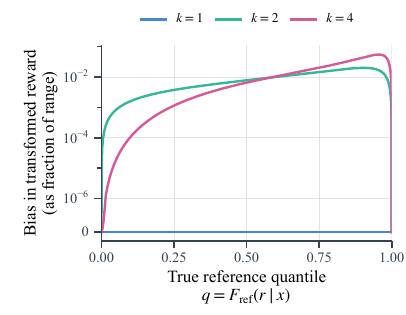}
        \caption{Bias in the transformed reward.}
    \end{subfigure}
    \caption{Win-rate transformation and exact finite-bank noise and bias for $M=30$ and $\beta=0.1$. To compare different $k$, values are normalized by $\Delta_T=T(1)-T(0)$; this normalization is only for display, and the transformation panel additionally subtracts $T(0)$. The horizontal axis is the true reference quantile of a fixed raw reward. Noise and bias use the same axis conventions as Figure~\ref{fig:calibration_binary_errors}.}
    \label{fig:calibration_winrate_errors}
\end{figure}

\paragraph{Sensitivity and convergence guarantees.}

\begin{lemma}[Consistency and error bounds for win-rate calibration]
\label{lem:calibration_winrate_guarantees}
Fix an integer $k\geq1$ and $\beta>0$. The transformation $T$ in Eq.~\ref{eq:calibration_winrate_map} is convex and $k$-Lipschitz. Under i.i.d. reference sampling, for every $q\in[0,1]$ and bank size $M$, the estimator $T(\widehat q)$ satisfies
\begin{equation}\label{eq:calibration_winrate_error_bounds}
\sigma_T(q)
\leq k\sqrt{\frac{q(1-q)}{M}},
\qquad
0\leq b_T(q)
\leq
\frac{k(k-1)(C+\beta^{-1})q(1-q)}{2M}.
\end{equation}
At fixed interior $q$, the asymptotic expressions in Eq.~\ref{eq:calibration_asymptotic_rates} apply with $\theta=T$. At $q=0$ and $q=1$, both errors are zero.

For each fixed prompt, the estimated mapping
$\widehat S_x^{\mathrm q}(r):=T(\widehat F_{\mathrm{ref}}(r\mid x))$
converges uniformly almost surely to $S_x^{\mathrm q}(r)$. Moreover, for every $\delta\in(0,1)$,
\begin{equation}\label{eq:calibration_winrate_uniform_bound}
\Pr\left(
\sup_r
\left|\widehat S_x^{\mathrm q}(r)-S_x^{\mathrm q}(r)\right|
>
k\sqrt{\frac{\log(2/\delta)}{2M}}
\right)
\leq\delta.
\end{equation}
These estimation guarantees hold for both continuous and discrete reference reward distributions.
\end{lemma}

\begin{proof}
\textit{Shape of the transformation.}
Using $G'(q)=C+G(q)^k/\beta$, differentiation gives, for $k>1$,
\begin{equation}\label{eq:calibration_winrate_derivatives}
T'(q)=kG(q)^{k-1},
\qquad
T''(q)=k(k-1)G(q)^{k-2}
\left(C+\frac{G(q)^k}{\beta}\right).
\end{equation}
Since $G(q)\in[0,1]$, these derivatives satisfy
\[
0\leq T'(q)\leq k,
\qquad
0\leq T''(q)\leq k(k-1)(C+\beta^{-1}).
\]
Thus, $T$ is convex and $k$-Lipschitz. For $k=1$, direct differentiation gives $T'(q)=1$ and $T''(q)=0$, so the same conclusions hold.

\textit{Pointwise noise and bias.}
The Lipschitz property gives
\[
\operatorname{Var}(T(\widehat q))
\leq
\mathbb E\left[(T(\widehat q)-T(q))^2\right]
\leq
k^2\mathbb E[(\widehat q-q)^2]
=
\frac{k^2q(1-q)}{M}.
\]
Taking square roots proves the noise bound. Convexity and $\mathbb E[\widehat q]=q$ give $b_T(q)\geq0$ by Jensen's inequality. A second-order Taylor expansion with the uniform bound on $T''$ gives
\[
T(\widehat q)
\leq
T(q)+T'(q)(\widehat q-q)
+\frac{k(k-1)(C+\beta^{-1})}{2}(\widehat q-q)^2.
\]
Taking expectations proves the upper bound on bias. At fixed interior $q$, the Taylor argument from Lemma~\ref{lem:calibration_binary_consistency} yields Eq.~\ref{eq:calibration_asymptotic_rates}. At either endpoint, $\widehat q=q$ almost surely, so both errors vanish.

\textit{Uniform consistency.}
The Lipschitz property also implies
\[
\sup_r
\left|\widehat S_x^{\mathrm q}(r)-S_x^{\mathrm q}(r)\right|
\leq
k\sup_r
\left|\widehat F_{\mathrm{ref}}(r\mid x)-F_{\mathrm{ref}}(r\mid x)\right|.
\]
Applying the Dvoretzky--Kiefer--Wolfowitz--Massart inequality~\citep{massart1990tight} proves Eq.~\ref{eq:calibration_winrate_uniform_bound}. More generally, for every $\varepsilon>0$,
\[
\Pr\left(
\sup_r
\left|\widehat S_x^{\mathrm q}(r)-S_x^{\mathrm q}(r)\right|
>\varepsilon
\right)
\leq
2\exp\left(-\frac{2M\varepsilon^2}{k^2}\right).
\]
The right-hand side is summable over $M$, so the Borel--Cantelli lemma gives uniform almost-sure convergence. Neither the binomial calculation nor the empirical-CDF inequality requires a continuous reference distribution.
\end{proof}

\paragraph{Where quantile error matters.}
Equation~\ref{eq:calibration_winrate_derivatives} shows that sensitivity increases toward the upper reference quantiles. Convexity also explains the upward bias: although the empirical CDF is unbiased at each fixed threshold, the transformed reward generally is not. For $k=1$, $T$ is affine, so the bias is zero and the standard deviation is exactly $\sqrt{q(1-q)/M}$.

The finite-sample bounds establish $O(M^{-1/2})$ noise and $O(M^{-1})$ bias for fixed $k$ and $\beta$, while Eq.~\ref{eq:calibration_asymptotic_rates} gives the leading constants at fixed interior quantiles. These guarantees concern estimation of the mapping $T(F_{\mathrm{ref}}(r\mid x))$; they do not require the reference win rates themselves to be uniformly distributed.

\paragraph{Implications.}
The win-rate transformation has the opposite sensitive region: both noise and bias are small at low and intermediate quantiles but increase toward the upper tail, where $T$ is steepest, before vanishing at $q=1$. At $k=4$, the standard deviation and bias of the transformed reward are $2.62\%$ and $0.66\%$ of its range at $q=0.5$, but increase to $18.27\%$ and $4.43\%$ at $q=0.9$. The bias is positive because the convex transformation amplifies upward empirical-CDF errors more than downward ones. Near the upper boundary, the main finite-sample event is assigning the maximal empirical quantile to a reward whose true quantile is below one; for example, at $q=0.99$, this occurs with probability $\Pr(\widehat q=1)=q^{30}\approx0.740$.

Both effects vanish as $M\to\infty$, as $\widehat q$ concentrates around $q$ and the probability $q^M$ of assigning a premature maximal quantile decreases. Additional reference completions are therefore most useful when accurate upper-tail ranks are important, particularly where $M(1-q)$ is small. Since the same reference bank is reused throughout training, its error is a fixed distortion of the estimated transformation rather than fresh noise at every update. Overall, $M=30$ provides substantially better resolution away from the sensitive boundaries, while larger banks can be targeted to hard binary prompts or to applications requiring more accurate upper-tail quantiles. For continuous rewards, this analysis isolates uncertainty from estimating the reference CDF; the separate approximation introduced by tied or discrete rewards is discussed in Appendix~\ref{app:winrate_maxk_derivation}.

\section{Comparison with Direct Multi-Completion Optimization}
\label{app:math_pkpo_comparison}

\paragraph{Checkpoint selection.}
For each model, method, and training value
$k_{\mathrm{train}}\in\{2,4\}$, we select the checkpoint that maximizes
$\mathrm{pass@k}_{\mathrm{train}}$ on the MATH Level~5 validation
problems. Figure~\ref{fig:math_pkpo_comparison}
reports the complete validation curves of the selected checkpoints, and Table~\ref{tab:math_pkpo_comparison} reports the corresponding numerical values.

At the evaluation budget matching the training objective,
$\mathrm{@k}$-$A^{\star}$PO is within $0.5$ percentage points of
PKPO in every comparison. The two methods also produce similar
performance across the remaining evaluation budgets. This supports
the central prediction of our analysis: adaptive pressure applied
through an ordinary single-completion optimizer can reproduce direct
multi-completion $\mathrm{pass@k}$ optimization.

\begin{figure}[t]
    \centering
    \begin{subfigure}[t]{0.492\linewidth}
        \centering
        \includegraphics[width=\linewidth]
        {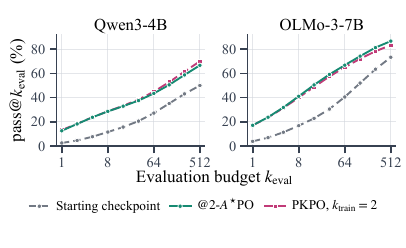}
        \caption{$k_{\mathrm{train}}=2$.}
        \label{fig:aime26_pkpo_k2}
    \end{subfigure}\hfill
    \begin{subfigure}[t]{0.492\linewidth}
        \centering
        \includegraphics[width=\linewidth]
        {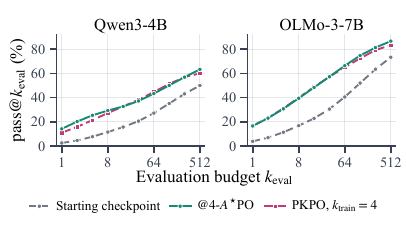}
        \caption{$k_{\mathrm{train}}=4$.}
        \label{fig:aime26_pkpo_k4}
    \end{subfigure}
    \caption{
    AIME 2026 test performance of the starting checkpoints
    and the selected $\mathrm{@}k$-$A^{\star}$PO and PKPO
    checkpoints across evaluation sampling budgets.
    Each subfigure reports Qwen3-4B (left) and OLMo-3-7B (right).
    }
    \label{fig:aime26_pkpo_comparison}
\end{figure}

Figure~\ref{fig:aime26_pkpo_comparison} extends this comparison to
AIME 2026. Across both model families and training values of $k$,
the two methods achieve comparable performance across evaluation
budgets, showing that their agreement extends beyond the
in-distribution validation set to harder, unseen problems.

\begin{table}[t]
\centering
\caption{
MATH Level~5 validation performance (in \%) of the starting checkpoints and selected $\mathrm{@k}$-$A^{\star}$PO and PKPO checkpoints, corresponding to Figure~\ref{fig:math_pkpo_comparison}.
}
\label{tab:math_pkpo_comparison}
\small
\begin{tabular}{llccccc}
\toprule
Model & Method
& $\mathrm{pass@}1$
& $\mathrm{pass@}2$
& $\mathrm{pass@}4$
& $\mathrm{pass@}8$
& $\mathrm{pass@}16$ \\
\midrule
OLMo-3-7B
& Starting checkpoint
& 37.38 & 50.94 & 63.16 & 73.29 & 81.80 \\
& $\mathrm{@}2$-$A^{\star}$PO
& 65.34 & 76.98 & 84.38 & 89.11 & 92.22 \\
& PKPO ($k_{\mathrm{train}}=2$)
& 66.08 & 77.38 & 84.73 & 89.46 & 92.52 \\
& $\mathrm{@}4$-$A^{\star}$PO
& 64.46 & 76.51 & 84.39 & 89.33 & 92.45 \\
& PKPO ($k_{\mathrm{train}}=4$)
& 64.31 & 76.18 & 83.92 & 89.04 & 92.45 \\
\midrule
Qwen3-4B
& Starting checkpoint
& 23.93 & 38.32 & 54.32 & 68.16 & 77.79 \\
& $\mathrm{@}2$-$A^{\star}$PO
& 57.27 & 70.48 & 79.39 & 85.34 & 89.43 \\
& PKPO ($k_{\mathrm{train}}=2$)
& 59.88 & 70.77 & 78.77 & 84.59 & 88.75 \\
& $\mathrm{@}4$-$A^{\star}$PO
& 57.22 & 70.14 & 78.63 & 84.22 & 88.29 \\
& PKPO ($k_{\mathrm{train}}=4$)
& 57.94 & 69.91 & 78.65 & 84.81 & 89.20 \\
\bottomrule
\end{tabular}
\end{table}

\paragraph{Rollout-compute comparison.}
With a prompt batch of 128 and 1,000 training steps,
$\mathrm{@k}$-$A^{\star}$PO generates $128{,}000$ on-policy
completions per run. PKPO generates $n$ completions for every prompt
and therefore requires $128{,}000n$ completions. At its minimum
rollout multiplicity $n=k$, this gives $256{,}000$ completions for
$k=2$ and $512{,}000$ for $k=4$. Thus,
$\mathrm{@k}$-$A^{\star}$PO reduces online rollout generation by
$50\%$ and $75\%$, respectively; larger PKPO groups increase this
difference proportionally.

$\mathrm{@k}$-$A^{\star}$PO additionally requires one-time reference
calibration, whose implementation and cost are detailed in
Appendix~\ref{app:reference_calibration}. In our setup, generating the
complete reference bank took approximately half an hour, compared with
$10$--$12$ hours for one training run, and the same bank is reused
across all configurations of a given model and dataset. Even when
amortized over only six runs, calibration therefore adds approximately
five minutes per run, or less than $1\%$ of training time. Its
contribution is consequently small relative to the savings from
removing PKPO's $k$-dependent online rollout multiplicity.

\end{document}